\documentclass[lettersize,journal]{IEEEtran}
\usepackage{amsmath,amsfonts}
\usepackage{algorithmic}
\usepackage{algorithm}
\usepackage{array}
\usepackage[caption=false,font=normalsize,labelfont=sf,textfont=sf]{subfig}
\usepackage{textcomp}
\usepackage{stfloats}
\usepackage{url}
\usepackage{verbatim}
\usepackage{graphicx}
\usepackage{multirow}
\usepackage{cite}
\usepackage{booktabs}
\usepackage{xcolor}
\usepackage{subcaption}
\usepackage{makecell}
\usepackage{marvosym}
\usepackage{listings}
\begin{document}

\title{PROBE: Diagnosing Residual Concept Capacity in Erased Text-to-Video Diffusion Models}

\author{
Yiwei Xie, Zheng Zhang\textsuperscript{~\Letter}, Ping Liu\textsuperscript{~\Letter}
% \textsuperscript{†} 
        % <-this % stops a space
\thanks{Yiwei Xie and Zheng Zhang are with the School of Artificial Intelligence and Automation, Huazhong University of Science and Technology, Wuhan 430074, China (email: \{yiweixie, leaf\}@hust.edu.cn)}% <-this % stops a space
\thanks{Ping Liu is with Department of Computer Science and Engineering, University of Nevada, Reno, NV, USA (email: pino.pingliu@gmail.com)}
\thanks{\textsuperscript{\Letter} denotes the co-corresponding author.}
% IEEE Publication Technology,~\IEEEmembership{Staff,~IEEE,}
        % <-this % stops a space
}



\maketitle

\begin{abstract}
Concept erasure techniques for text-to-video (T2V) diffusion models report substantial suppression of sensitive content, yet current evaluation is limited to checking whether the target concept is absent from generated frames, treating output-level suppression as evidence of representational removal. We introduce PROBE, a diagnostic protocol that quantifies the \textit{reactivation potential} of erased concepts in T2V models. 
With all model parameters frozen, PROBE optimizes a lightweight pseudo-token embedding through a denoising reconstruction objective combined with a novel latent alignment constraint that anchors recovery to the spatiotemporal structure of the original concept. 
We make three contributions: (1) a multi-level evaluation framework spanning classifier-based detection, semantic similarity, temporal reactivation analysis, and human validation; (2) systematic experiments across three T2V architectures, three concept categories, and three erasure strategies revealing that all tested methods leave measurable residual capacity whose robustness correlates with intervention depth; and (3) the identification of temporal re-emergence, a video-specific failure mode where suppressed concepts progressively resurface across frames, invisible to frame-level metrics.
These findings suggest that current erasure methods achieve output-level suppression rather than representational removal. 
We release our protocol to support reproducible safety auditing.
Our code is available at https://github.com/YiweiXie/PRObingBasedEvaluation.
\end{abstract}


\section{Introduction}
\label{sec:introduction}

\IEEEPARstart{C}{oncept} erasure techniques for text-to-video (T2V) diffusion models aim to suppress the generation of sensitive, prohibited, or copyrighted content without full retraining~\cite{xie2025erasing_arxiv2025}. 
Methods adapted from the text-to-image (T2I) domain, including negative prompting~\cite{li2024negprompt}, inference-time activation steering~\cite{yoon2025safree}, and parameter-efficient unlearning~\cite{gandikota2023esd}, report substantial reductions in the visible occurrence of targeted concepts. 
However, erasure success is typically judged by a single criterion: whether the target concept is absent from generated outputs~\cite{pham2023circumventing_nips2024,lu2025concepts_arxiv2025}. 
This conflates two different outcomes, namely surface-level suppression and representational deletion, leaving a critical question unaddressed:

\medskip
\noindent\textit{Do current T2V erasure methods truly remove target concepts from the model's internal representations, or do they merely suppress their visible occurrence under standard prompting?}
\medskip

This question is particularly pressing in the video setting. 
Unlike static images, videos distribute semantics across temporal sequences: a concept suppressed in early frames may re-emerge later as latent activations propagate through temporal attention layers~\cite{chefer2025videojam_icml2025}. 
Modern T2V architectures such as CogVideoX~\cite{yang2025cogvideox} and Wan~\cite{wang2025wan_arxiv2025} couple spatial and temporal dimensions through shared attention mechanisms. 
As a result, concept traces that survive erasure in the spatial representation can propagate through temporal attention layers, manifesting as delayed or progressive reactivation patterns across the generated sequence. 
Frame-level evaluation, which dominates current practice, is structurally unable to detect such temporal failure modes.

This concern is not hypothetical. 
In the T2I domain \cite{jiacheng2024text-guide}, concepts declared ``erased'' have been recovered through pseudo-token optimization~\cite{pham2023circumventing_nips2024,tsai2024ringabell} and adversarial prompt construction~\cite{beerens2025vulnerability_arxiv2025}, demonstrating that output-level suppression does not guarantee representational removal. 
Whether comparable vulnerabilities extend to T2V models, where temporal dynamics introduce additional recovery pathways, has not been systematically investigated despite growing interest in video-domain erasure~\cite{liu2024unlearning_arxiv2024, xu2025videoeraser_emnlp2025, ye2025t2vunlearning_arxiv2025, yoon2025safree}.

\begin{figure*}[!t]
\centering
\includegraphics[width=5.5 in]{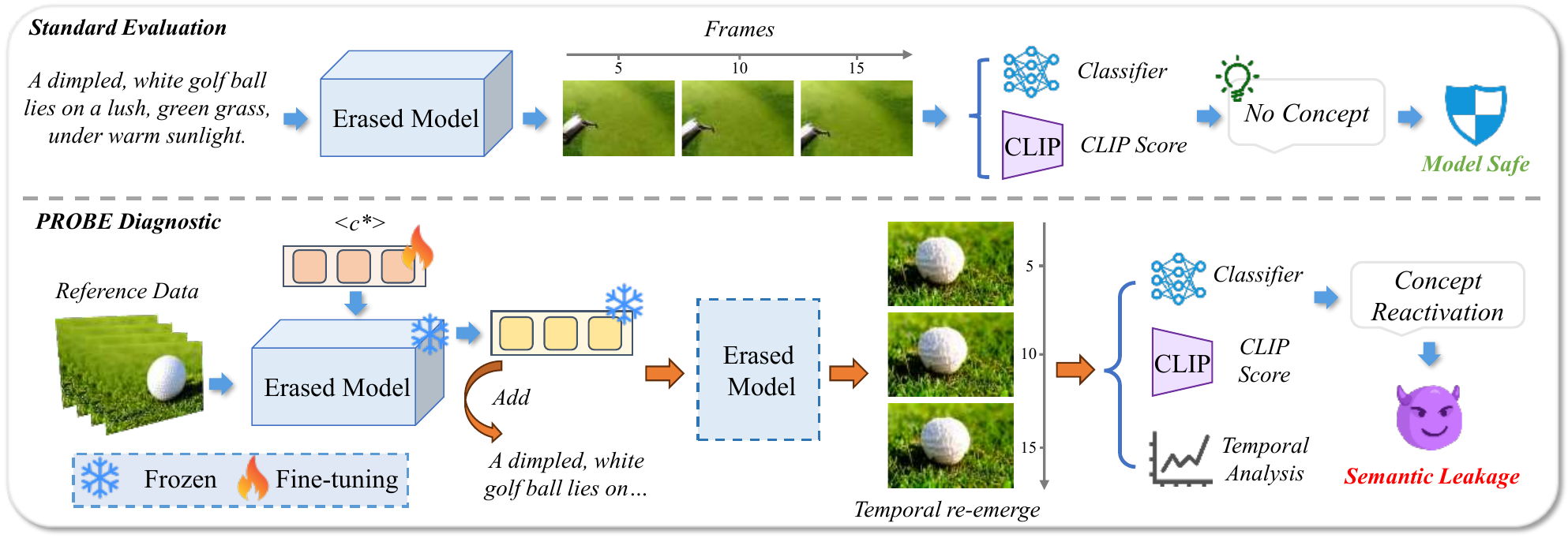}
\caption{Overview of the PROBE diagnostic framework.}
\label{fig:PROBE Framework}
\end{figure*}

% \IEEEpubidadjcol
To address this gap, we introduce PROBE (PRObing-Based Evaluation), a diagnostic protocol that tests whether erased T2V models still retain the capacity to regenerate a target concept under controlled recovery. 
We term this residual recoverability the \textit{reactivation potential} of an erased concept.
As illustrated in Fig.~\ref{fig:PROBE Framework}, PROBE probes residual capacity through the narrowest possible intervention: optimizing only a pseudo-token embedding while keeping all model parameters frozen.
This design is motivated by a diagnostic principle: because no model weights are modified, any successful recovery must originate from information already encoded in the frozen parameters, not from new knowledge introduced during probing.
The approach shares the optimization mechanics of textual inversion~\cite{gal2023TextInversion_iclr2023}, which has been widely adopted for concept 
customization~\cite{jiang2024animediff_tmm2024,xu2024sgdm_tmm2024} but repurposes it from personalization to diagnostic measurement, with an additional latent alignment objective (Sec.~\ref{sec:alignment}) that anchors recovery to the spatiotemporal structure of the original concept.

We apply PROBE across three T2V architectures (CogVideoX-2B, CogVideoX-5B, Wan2.2-5B), three concept categories (objects, NSFW content, and celebrity identities), and erasure strategies spanning input conditioning, activation steering, and weight-space unlearning. 
Our experiments reveal four principal findings. 
First, all tested erasure methods leave measurable residual capacity that can be reactivated under controlled probing. 
Second, erasure robustness correlates with intervention depth: input-level methods are most vulnerable, while weight-space unlearning provides stronger but still incomplete removal. 
Third, temporal analysis reveals delayed re-emergence patterns, where concepts suppressed in early frames progressively resurface across the generated sequence, that are invisible to frame-level metrics. 
Fourth, learned probes transfer across erasure methods within the same model family, indicating that residual capacity is partially shared across erasure paradigms.

Our contributions are summarized as follows:
\begin{itemize}
    \item We introduce a diagnostic protocol for T2V concept erasure evaluation that integrates explicit threat model assumptions, embedding-level probing with frozen model parameters, and a standardized multi-level evaluation framework.
    \item We define \textit{reactivation potential} as a quantitative measure of erasure robustness and propose a dual-objective probing mechanism combining diagnostic reconstruction with latent alignment to estimate it.
    \item We identify temporal re-emergence as a video-specific failure mode of concept erasure, where suppressed concepts progressively resurface across generated frames, and provide temporal reactivation curves as a diagnostic tool to detect it.
    \item We conduct a systematic empirical study across three architectures, three concept categories, and three erasure strategies, demonstrating that no tested method achieves complete representational removal. We release our protocol and implementation to support reproducible safety auditing.
\end{itemize}

% \begin{figure*}[!t]
% \centering
% \includegraphics[width=\textwidth]{motivation_figure.pdf}
% \caption{Comparison of standard evaluation and PROBE diagnostic for concept erasure in T2V models. 
% (a)~Standard practice checks only whether the target concept is absent from generated frames, declaring erasure successful when no concept is detected under direct prompting. 
% (b)~PROBE optimizes a pseudo-token embedding against the frozen erased model, revealing residual concept capacity that manifests as temporal re-emergence: the target concept is suppressed in early frames but progressively resurfaces across the generated sequence (indicated by intensifying frame colors), a failure mode invisible to frame-level evaluation.}
% \label{fig:motivation}
% \end{figure*}

\section{Related Work}
\label{sec:related}

\subsection{Text-to-Video Diffusion Models}

Early T2V systems adapt U-Net backbones with temporal attention layers~\cite{ho2022video_nips2022,wang2023VideoComposer_nips2023,wu2023tune_iccv2023}, while more recent architectures transition toward Diffusion Transformers (DiT)~\cite{peebles2023scalable_iccv2023, menapace2024_snapvideo_CVPR, mao2026LDT_ieeeTMM2026} that jointly attend over text and video tokens.
CogVideoX~\cite{yang2025cogvideox} introduces multimodal DiTs with v-prediction parameterization, and Wan~\cite{wang2025wan_arxiv2025} adopts flow matching with a mixture-of-experts design.
Other notable systems include HunyuanVideo~\cite{kong2024hunyuanvideo} and Open-Sora~\cite{zheng2024open_arxiv2024}.
These advances have broadened T2V applications across controllable generation~\cite{hu2023benchmark_tmm2023, li_starvid_tmm2026}, audio-visual synthesis~\cite{zhao2024ta2v_tmm2024, Ishii_CMC-PE_ijcnn2025}, and subject-driven customization~\cite{chen2025videodreamer_tmm2025, wang2026_CustomVideo_TMM2026}, but the same representational capacity also allows T2V models to produce unsafe content~\cite{xie2025erasing_arxiv2025,nie2024dip_tmm2024, pan2025_thesafetyillusion_tip2025}, motivating concept erasure research in the video domain.

\subsection{Concept Erasure in Generative Models}
In T2I, erasure methods span input-level conditioning~\cite{sridhar2024prompt_eccv2024}, activation-level steering~\cite{lu2024mace_cvpr2024,Li2025ANT_acmmm2025, lee2025_ESC_cvpr2025}, and parameter-level editing~\cite{chavhan2025conceptprune_iclr2025,huang2024receler_eccv2024,gandikota2024uce_wacv2024,lyu2024one_cvpr2024}, with recent extensions to Flux-based~\cite{gao2024eraseanything} and autoregressive models~\cite{fan2025earerasingconceptsunified_arxiv2025}; we refer readers to~\cite{xie2025erasing_arxiv2025} for a comprehensive survey.
T2V erasure methods mirror this taxonomy but remain limited in number.
NegPrompt~\cite{li2024negprompt} suppresses target concepts at the input level, SAFREE~\cite{yoon2024safree_arxiv2024} steers cross-attention at the activation level, and T2VUnlearning~\cite{ye2025t2vunlearning_arxiv2025} applies LoRA-based fine-tuning at the parameter level.
Other approaches include training-free embedding adjustments~\cite{xu2025videoeraser_emnlp2025} and loss-based text encoder modification~\cite{liu2025erased_arxiv2025}.
However, evaluation practices vary substantially across studies in metrics, concept categories, and generation configurations, with none including adversarial recovery testing.

\subsection{Robustness Evaluation of Concept Erasure}

In the T2I domain, erased concepts have been recovered through pseudo-token optimization~\cite{pham2024conceptinvention_iclr2024} and adversarial prompt search~\cite{tsai2024ringabell,chin2024p4d, yang_CAAM_icme2025}, while adversarial defense methods attempt to improve robustness through joint training~\cite{zhang2024defensive_nips2024} or multi-modal attacks~\cite{beerens2025vulnerability_arxiv2025}.
Analytical studies examine theoretical conditions for erasure completeness~\cite{lu2025concepts_arxiv2025} and show that erased concepts may remain dormant rather than permanently removed~\cite{liu2025erased_arxiv2025}.
EMMA~\cite{wei2025emma_arxiv2025} provides multi-dimensional benchmarking for T2I erasure but does not include adversarial recovery testing.
These studies collectively establish that output-level suppression does not imply representational removal.
% in T2I models.

Robustness evaluation for T2V erasure remains unexplored.
Existing T2V methods rely on frame-level classifiers and direct prompting~\cite{xu2025videoeraser_emnlp2025,ye2025t2vunlearning_arxiv2025,yoon2024safree_arxiv2024}, which cannot capture video-specific failure modes such as temporal re-emergence.
The closest methodological precedents, pseudo-token 
optimization~\cite{pham2024conceptinvention_iclr2024} and 
Ring-A-Bell~\cite{tsai2024ringabell}, differ from PROBE in three 
respects: they target static images and do not address temporal 
re-emergence as a failure mode, they use reconstruction-only 
objectives without latent alignment, and they lack a multi-level 
evaluation protocol.
To our knowledge, no diagnostic protocol has been proposed for assessing concept recoverability in the video domain.

\section{Preliminaries}
\label{sec:preliminaries}

This section introduces notation for text-to-video diffusion models and textual inversion, which form the technical foundation of our diagnostic protocol.

\subsection{Text-to-Video Diffusion Models}

A T2V diffusion model operates on spatiotemporal latents $\mathbf{x}_0 \in \mathbb{R}^{T \times H \times W \times C}$, where $T$ denotes the number of frames and $H, W, C$ denote spatial height, width, and channel dimensions. Given a text condition $c$, the model learns a denoising network that jointly models spatial content and temporal dynamics. We support two prevalent parameterizations.

\vspace{4pt}
\noindent\textbf{V-prediction.}
Models such as CogVideoX~\cite{yang2025cogvideox} predict a velocity field $\mathbf{v}_\theta(\mathbf{x}_t, c, t)$ from a noisy latent $\mathbf{x}_t = \alpha_t \mathbf{x}_0 + \sigma_t \boldsymbol{\epsilon}$, where $\boldsymbol{\epsilon} \sim \mathcal{N}(\mathbf{0}, \mathbf{I})$. The clean latent is recovered via:
\begin{equation}
    \hat{\mathbf{x}}_0 = \sqrt{\alpha_t} \mathbf{x}_t - \sqrt{1 - \alpha_t} \cdot \mathbf{v}_\theta(\mathbf{x}_t, c, t).
    \label{eq:vpred}
\end{equation}

\vspace{4pt}
\noindent\textbf{Flow matching.}
Models such as Wan2.2~\cite{wang2025wan_arxiv2025} learn a continuous-time velocity field under a linear interpolation path $\mathbf{x}_t = (1 - \sigma_t)\boldsymbol{\epsilon} + \sigma_t \mathbf{x}_0$. The clean latent is inferred via:
\begin{equation}
    \hat{\mathbf{x}}_0 = \mathbf{x}_t - t \cdot \mathbf{v}_\theta(\mathbf{x}_t, c, t).
    \label{eq:flowmatch}
\end{equation}

\noindent Both formulations share a common structure: a parameterized velocity field $\mathbf{v}_\theta$ conditioned on text input $c$, from which a clean latent prediction $\hat{\mathbf{x}}_0$ can be derived at each denoising step. This shared structure allows our protocol to operate uniformly across both paradigms.

\subsection{Textual Inversion}

Textual inversion~\cite{gal2023TextInversion_iclr2023} and subsequent personalization 
methods~\cite{xu2024sgdm_tmm2024} capture a visual concept by optimizing a learnable pseudo-token embedding $\mathbf{v} \in \mathbb{R}^d$ within the text encoder's embedding space, while all model parameters $\theta$ remain frozen:
\begin{equation}
    \mathcal{L}_{\mathrm{TI}}(\mathbf{v}) = \mathbb{E}_{\mathbf{x}_0, t, \boldsymbol{\epsilon}} \left[ \left\| \boldsymbol{\epsilon} - \boldsymbol{\epsilon}_\theta(\mathbf{x}_t, \mathbf{v}, t) \right\|_2^2 \right].
    \label{eq:textual_inversion}
\end{equation}
The optimized embedding $\mathbf{v}$ can then be inserted into arbitrary text prompts to reproduce the learned concept. 
Our protocol builds on this mechanism but repurposes it for diagnostic measurement with an additional latent alignment objective designed for the spatiotemporal setting of T2V models (Sec.~\ref{sec:alignment}).

\section{Probing Residual Concept Capacity in Erased T2V Models}
\label{sec:protocol}

This section presents the PROBE protocol.
We begin by specifying the access assumptions that constrain probing to a single learnable pseudo-token embedding (Sec.~\ref{sec:threat}).
Within this constraint, we define a diagnostic reconstruction loss that tests whether the frozen model can still be steered toward the target concept (Sec.~\ref{sec:reconstruction}).
Because reconstruction alone may converge to co-occurring attributes rather than the concept itself, we further propose a latent alignment loss that anchors recovery to the spatiotemporal structure of the original concept, providing a key distinction from standard textual inversion (Sec.~\ref{sec:alignment}).
We then describe the multi-level evaluation framework that quantifies recovery across complementary dimensions (Sec.~\ref{sec:evaluation}), followed by a temporal reactivation analysis that captures video-specific failure modes invisible to frame-level metrics (Sec.~\ref{sec:temporal}).
The complete protocol is formalized in Algorithm~\ref{alg:probe}.

% --------------------------------------------------------------
\subsection{Access Assumptions and Constraints}
\label{sec:threat}

PROBE is designed to measure what remains inside an erased model, not to introduce new information into it. This diagnostic principle dictates three constraints.
First, all model parameters remain strictly frozen; the only learnable component is a pseudo-token embedding $\mathbf{v} \in \mathbb{R}^d$ optimized within the text encoder's embedding space. This ensures that any successful recovery must originate from information already encoded in the frozen parameters, not from capacity introduced during probing.
Second, the protocol assumes no knowledge of the erasure method's loss functions, training procedures, or architectural modifications, requiring only forward access to the erased model's text encoder and denoising network. This makes the diagnostic applicable to any erasure method without method-specific adaptation.
Third, a small set of reference videos depicting the target concept $c^*$ is required as optimization targets. These are generated from the unerased model using prompts containing $c^*$, rather than collected from external sources, to standardize the reference distribution and isolate the measurement from data-quality confounds. We analyze the sensitivity to reference set size in Sec.~\ref{sec:ablation}.

The evaluation setting is intentionally restrictive: if PROBE can reactivate a concept under these conditions, the erasure is demonstrably incomplete. Conversely, failure to recover does not prove complete removal, but indicates robustness against embedding-level probing.
Within these constraints, the probing procedure optimizes the pseudo-token embedding through two complementary objectives: diagnostic reconstruction (Sec.~\ref{sec:reconstruction}) and latent alignment (Sec.~\ref{sec:alignment}).

% --------------------------------------------------------------
\subsection{Diagnostic Reconstruction}
\label{sec:reconstruction}

The first objective adapts the textual inversion loss~\cite{gal2023TextInversion_iclr2023} to the erased model, testing whether it can still be steered toward the target concept through embedding-level optimization.
Given an erased T2V model with frozen parameters $\theta'$ and reference clips with latent encodings $\mathbf{z}_0^{(i)}$, we construct noised latents $\mathbf{z}_t^{(i)} = \alpha_t \mathbf{z}_0^{(i)} + \sigma_t \boldsymbol{\epsilon}$ and optimize a shared pseudo-token embedding $\mathbf{v}$ using the denoising objective:
\begin{equation}
    \mathcal{L}_{\text{rec}}(\mathbf{v}) = \mathbb{E}_{i, t, \boldsymbol{\epsilon}} \left[ \left\| \boldsymbol{\epsilon} - \boldsymbol{\epsilon}_{\theta'}(\mathbf{z}_t^{(i)}, \mathbf{v}, t) \right\|_2^2 \right].
    \label{eq:rec}
\end{equation}

The embedding $\mathbf{v}$ is shared across all frames and diffusion timesteps, so its gradients are aggregated over the entire spatiotemporal generation trajectory. 
As a result, any reduction in the denoising loss must reflect a representation that is coherent across the full video sequence, rather than relying on frame-specific shortcuts.

% --------------------------------------------------------------
\subsection{Latent Alignment}
\label{sec:alignment}

The reconstruction objective alone does not guarantee that recovered content corresponds to the target concept.
Consider erasing the concept ``dog'' from a T2V model. 
When we optimize a pseudo-token using only $\mathcal{L}_{\text{rec}}$, the embedding may converge to a representation that reduces denoising error by reproducing surface-level attributes frequently co-occurring with dogs in the training data, such as grass, leashes, or outdoor scenes, without actually recovering the dog itself. 
This co-occurrence phenomenon, where models exploit spurious correlations between target concepts and co-occurring context rather than learning the concept itself, is well-documented in visual recognition~\cite{11330177_tmm2026}.
This is because $\mathcal{L}_{\text{rec}}$ operates in the noise space, where semantically distinct outputs can yield similar denoising errors. Our ablation study (Sec.~\ref{sec:ablation}) confirms this effect: removing the alignment term consistently reduces recovery quality across all tested configurations.

We address this by introducing a latent alignment objective that operates in the clean latent space, where co-occurring attributes and the target concept occupy distinct representations. By directly comparing the model's predicted clean latent against reference latents of the target concept, the alignment loss penalizes embeddings that reduce denoising error through co-occurring cues while failing to recover the actual spatiotemporal structure of the concept.
Formally, let $\mathbf{z}_0^{(i)}$ denote the reference clean latent for clip $i$, and let $\hat{\mathbf{z}}_0^{(i)}(\mathbf{v}, t)$ denote the clean latent predicted by the frozen model at timestep $t$ when conditioned on the optimized embedding $\mathbf{v}$.
The alignment loss is defined as:
\begin{equation}
    \mathcal{L}_{\text{align}}(\mathbf{v}) = \mathbb{E}_{i, t} \left[ \left\| \hat{\mathbf{z}}_0^{(i)}(\mathbf{v}, t) - \mathbf{z}_0^{(i)} \right\|_2^2 \right].
    \label{eq:align}
\end{equation}

The predicted clean latent $\hat{\mathbf{z}}_0^{(i)}(\mathbf{v}, t)$ is computed from a single forward pass using the recovery formulas defined in Eqs.~(\ref{eq:vpred}) and~(\ref{eq:flowmatch}) for v-prediction and flow matching respectively.

The overall probing objective combines diagnostic reconstruction with latent alignment:
\begin{equation}
    \mathcal{L}_{\text{total}}(\mathbf{v}) = \mathcal{L}_{\text{rec}}(\mathbf{v}) + \lambda \, \mathcal{L}_{\text{align}}(\mathbf{v}),
    \label{eq:total}
\end{equation}
where $\lambda$ controls the strength of alignment. We analyze the sensitivity to $\lambda$ in Sec.~\ref{sec:ablation}.

% --------------------------------------------------------------
\subsection{Multi-Level Evaluation Framework}
\label{sec:evaluation}

The optimization procedure described above produces a pseudo-token embedding that may or may not recover the target concept. 
To assess whether recovery is genuine, we evaluate across multiple complementary levels, as no single metric can reliably distinguish concept recovery from detector artifacts.
For each metric, we define a standardized three-point comparison: the baseline generation rate of the unerased model, the post-erasure rate under direct prompting, and the post-PROBE rate after diagnostic probing (original $\rightarrow$ erased $\rightarrow$ probed).

\subsubsection{Classifier-Based Detection}
Classifier-based metrics provide the primary quantitative signal for concept recovery. Since different concept categories require different detection capabilities, we employ domain-specific classifiers following established practice~\cite{gandikota2023esd, lu2024mace_cvpr2024}. For object concepts, we report Top-1 and Top-5 classification accuracy via a pretrained ResNet-50~\cite{he2016resnet_cvpr2016}. For NSFW concepts, we use NudeNet~\cite{nudenet} detection rate, which provides fine-grained subclass predictions enabling analysis of which semantic components resist erasure. For celebrity identities, we compute cosine similarity between face recognition embeddings~\cite{deng2019arcface_cvpr2019}, referred to as ID-similarity.

\subsubsection{Semantic Similarity}
Classifier-based detection is inherently domain-specific and may exhibit biases tied to training distributions. To provide a detector-independent assessment, we measure CLIP-based similarity~\cite{radford2021clip_icml2021} between generated frames and text descriptions of the target concept $c^*$.
Formally, for each generated frame $I_j$ and a text description $p_{c^*}$ of the target concept, we compute:
\begin{equation}
    S_{\text{CLIP}} = \frac{1}{|\mathcal{F}|} \sum_{j \in \mathcal{F}} \text{sim}\!\left( f_{\text{img}}(I_j),\, f_{\text{txt}}(p_{c^*}) \right),
    \label{eq:clip}
\end{equation}
where $f_{\text{img}}$ and $f_{\text{txt}}$ denote the CLIP image and text encoders, and $\text{sim}(\cdot, \cdot)$ denotes cosine similarity. 
A higher $S_{\text{CLIP}}$ after PROBE relative to the post-erasure baseline indicates that the recovered content aligns semantically with the original concept.

\subsubsection{Human Validation}
Automated metrics may be susceptible to detector artifacts or distribution mismatch between generated and real content. 
As a final consistency check, we conduct human evaluation on a subset of generated videos.
Annotators provide binary judgments on whether the target concept is recognizably present, with inter-annotator agreement reported via Fleiss' $\kappa$. 
Agreement between automated and human assessments strengthens confidence that measured recovery reflects genuine concept restoration. 
Specific evaluation configurations are reported in Sec.~\ref{sec:experiments}.

% --------------------------------------------------------------
\subsection{Temporal Reactivation Analysis}
\label{sec:temporal}

The evaluation metrics described above all operate at the frame level and therefore cannot capture failure modes specific to video generation.
We identify one such failure mode, \textit{temporal re-emergence}: a concept that appears suppressed in early frames may progressively resurface across the generated sequence as latent activations propagate through temporal attention layers.
This phenomenon has no counterpart in T2I evaluation and, to our knowledge, has not been previously characterized in the concept erasure literature.

To detect temporal re-emergence, we compute the concept detection score at each frame and average across evaluation samples, producing a temporal reactivation curve. 
Formally, let $d(I_j^{(k)})$ denote the detection score for frame $j$ of generated video $k$. 
The temporal reactivation curve is defined as: 
\begin{equation}
    R(j) = \frac{1}{K} \sum_{k=1}^{K} d(I_j^{(k)}),
    \label{eq:temporal}
\end{equation}
where $K$ is the number of evaluation videos. 
The instantiation of the frame-level score $d(\cdot)$ is tailored to the concept category: for objects, $d(I_j)$ adopts a Top-1 confidence gating mechanism that yields the confidence score when the concept is within the Top-1 predictions and $0$ otherwise; for NSFW concepts, $d(I_j)$ uses the raw maximum detection probability to capture latent temporal risks; for identity concepts, $d(I_j)$ denotes the frame-level CLIP similarity.

We compare $R(j)$ across the three-point comparison (original, erased, and probed) to identify three temporal signatures. The first is \textbf{delayed onset}, where the concept is absent in early frames but reappears after a threshold frame index. The second is \textbf{mid-sequence resurgence}, where the concept is suppressed at the beginning and end but peaks in the middle. The third is \textbf{progressive accumulation}, where detection scores increase monotonically across frames, suggesting that temporal attention amplifies residual signals over time.
These patterns reveal how different erasure strategies interact with the model's temporal processing and provide diagnostic insights that frame-averaged metrics cannot capture.

% --------------------------------------------------------------
% Algorithm Box
% --------------------------------------------------------------
\begin{algorithm}[t]
\caption{PROBE: Diagnostic Protocol}
\label{alg:probe}
\scriptsize
\begin{algorithmic}[1]
\REQUIRE Erased model $\theta'$ (frozen), target concept $c^*$, reference videos $\mathcal{R} = \{V^{(i)}\}_{i=1}^{N}$
\ENSURE Reactivation potential across multi-level evaluation

\STATE \textbf{// Phase 1: Setup}
\STATE Encode reference videos: $\mathbf{z}_0^{(i)} \leftarrow \mathcal{E}(V^{(i)})$ for all $i$
\STATE Initialize pseudo-token $\mathbf{v}$ from $c^*$'s token embedding

\STATE \textbf{// Phase 2: Optimization}
\FOR{step $= 1$ to $S_{\max}$}
    \STATE Sample reference clip $i$, timestep $t$, noise $\boldsymbol{\epsilon}$
    \STATE Construct $\mathbf{z}_t^{(i)} = \alpha_t \mathbf{z}_0^{(i)} + \sigma_t \boldsymbol{\epsilon}$
    \STATE Compute $\mathcal{L}_{\text{total}}(\mathbf{v}) = \mathcal{L}_{\text{rec}}(\mathbf{v}) + \lambda \, \mathcal{L}_{\text{align}}(\mathbf{v})$ \hfill \emph{Eqs.~(\ref{eq:rec})--(\ref{eq:total})}
    \STATE Update $\mathbf{v} \leftarrow \mathbf{v} - \eta \nabla_{\mathbf{v}} \mathcal{L}_{\text{total}}$
\ENDFOR

\STATE \textbf{// Phase 3: Generation}
\STATE Insert optimized $\langle v^* \rangle$ into evaluation prompts
\STATE Generate videos $\{\hat{V}^{(k)}\}_{k=1}^{K}$ using erased model $\theta'$

\STATE \textbf{// Phase 4: Multi-Level Evaluation}
\STATE Compute classifier-based detection \hfill \emph{Sec.~\ref{sec:evaluation}}
\STATE Compute CLIP semantic similarity $S_{\text{CLIP}}$ \hfill \emph{Eq.~(\ref{eq:clip})}
\STATE Compute temporal reactivation curves $R(j)$ \hfill \emph{Sec.~\ref{sec:temporal}}
\STATE Conduct human validation with Fleiss' $\kappa$ \hfill \emph{Sec.~\ref{sec:evaluation}}

\STATE \textbf{Return} three-point comparison (original $\rightarrow$ erased $\rightarrow$ probed)
\end{algorithmic}
\end{algorithm}

\section{Experiments}
\label{sec:experiments}

We apply the PROBE protocol across three T2V architectures, three erasure strategies, and three concept categories. This section describes the experimental setup~(Sec.~\ref{sec:setup}), then presents residual capacity measurements for object concepts~(Sec.~\ref{sec:exp_object}), NSFW concepts~(Sec.~\ref{sec:exp_nsfw}), and celebrity identities~(Sec.~\ref{sec:exp_celeb}). We further report comparisons with adversarial baselines~(Sec.~\ref{sec:exp_baseline}), cross-model and cross-method transferability~(Sec.~\ref{sec:exp_transfer}), visual quality assessment~(Sec.~\ref{sec:exp_quality}), and ablation studies~(Sec.~\ref{sec:ablation}).

% --------------------------------------------------------------
\subsection{Experimental Setup}
\label{sec:setup}

\noindent\textbf{Models.}
We evaluate PROBE on three T2V architectures: CogVideoX-2B and CogVideoX-5B~\cite{yang2025cogvideox}, which adopt v-prediction with spatiotemporal attention, and Wan2.2-5B~\cite{wang2025wan_arxiv2025}, which uses flow matching with a distinct text encoder. The CogVideoX family enables controlled comparison across model scale, while Wan2.2 provides a complementary backbone with different inductive biases.

\noindent\textbf{Erasure methods.}
We evaluate three strategies spanning different intervention levels: NegPrompt~\cite{li2024negprompt}, an input conditioning method that appends negative tokens during inference without model modification; SAFREE~\cite{yoon2025safree}, an activation steering method that modifies cross-attention to de-weight concept-relevant features while keeping parameters frozen; and T2VUnlearning~\cite{ye2025t2vunlearning_arxiv2025}, a weight-space method that permanently modifies model parameters through LoRA-based fine-tuning~\cite{hu2022lora}.

\noindent\textbf{Concept categories.}
We consider three categories: everyday object concepts (\emph{e.g.}, ``garbage truck,'' ``church'') for evaluating fine-grained semantic forgetting, nudity-related NSFW content for safety-critical evaluation, and celebrity identities (Barack Obama and Donald Trump) for privacy-sensitive evaluation.

\noindent\textbf{Evaluation metrics.}
Following the multi-level framework in Sec.~\ref{sec:evaluation}, we report classifier-based metrics (Top-1/Top-5 via ResNet-50 for objects, NudeNet~\cite{nudenet} for NSFW, ID-similarity via ArcFace~\cite{deng2019arcface_cvpr2019} for identities), CLIP-based similarity~\cite{radford2021clip_icml2021} as a detector-independent measure, temporal reactivation curves as defined in Sec.~\ref{sec:temporal}, and human validation with Fleiss' $\kappa$ for inter-annotator agreement. 
For human validation, annotators classify generated outputs in a blind setup, aggregated via majority vote. 
For each configuration, we report the three-point comparison: original, post-erasure, and post-PROBE.

% \noindent\textbf{Evaluation metrics.}
% Following the multi-level framework in Sec.~\ref{sec:evaluation}, we report classifier-based metrics (Top-1/Top-5 via ResNet-50 for objects, NudeNet~\cite{nudenet} for NSFW, ID-similarity via ArcFace~\cite{deng2019arcface_cvpr2019} for identities), CLIP-based similarity~\cite{radford2021clip_icml2021} as a detector-independent measure, temporal reactivation curves as defined in Sec.~\ref{sec:temporal}, and human validation with Fleiss' $\kappa$ for inter-annotator agreement. 
% For human validation, annotators classify generated outputs in a blind setup, and preferences are aggregated via majority vote. 
% For each configuration, we report the three-point comparison: original, post-erasure, and post-PROBE.

\noindent\textbf{Implementation details.}
\label{sec:implementationdetails}
The pseudo-token embedding $\mathbf{v}$ is initialized from the target concept's original token embedding when available, or from the mean of common neutral descriptors otherwise. During optimization, prompt augmentation is applied by randomly paraphrasing the reference prompt at each step. We use AdamW~\cite{adamw_iclr2019} with cosine learning rate decay, warmup, and gradient clipping. The default configuration uses 5 pseudo-tokens, a learning rate of 0.02, and $\lambda = 1$. Training typically converges within 1k--3k steps.

We generate 100 reference videos for nudity concepts, 20 per object category, and 30 per celebrity identity. All experiments use a single NVIDIA H100 GPU with a fixed random seed of 42. For CogVideoX-2B, we use 49, 17, and 17 frames per video for nudity, object, and celebrity concepts respectively, with training times of approximately 10, 3, and 3 hours. For CogVideoX-5B and Wan2.2-5B, we use 25 frames per video, taking approximately 7 hours per category. During generation, we use CFG\,=\,6.0 for CogVideoX and CFG\,=\,5.0 for Wan2.2, with 50 sampling steps. Spatial resolutions are 720$\times$480 for CogVideoX and 1280$\times$720 for Wan2.2. 
Our modular codebase supports all three evaluated 
erasure methods and provides extensible interfaces 
for integrating future mechanisms. 
More details are provided in supplementary material.

% In most cases, the default 5 pseudo-tokens lead to stable convergence. For certain object categories, shorter representations yield more stable optimization; concept-specific adjustments are detailed in the supplementary material.

% ==============================================================
\subsection{Residual Capacity on Object Concepts}
\label{sec:exp_object}

We begin by evaluating residual concept capacity on object categories in CogVideoX-2B, which represent a broad class of everyday semantics beyond safety-critical settings.

\begin{table*}[t]
\centering
\caption{Object erasure results on CogVideoX-2B across three metrics (Erased / After PROBE). T1: Top-1 accuracy (\%), T5: Top-5 accuracy (\%), CL: CLIP similarity. The highest post-PROBE recovery per row per metric is in \textbf{bold}.}
\label{tab:object_all}
\scriptsize
\setlength{\tabcolsep}{1.5pt}
\begin{tabular}{l ccc ccc ccc ccc}
\toprule
& \multicolumn{3}{c}{\bf{Origin}} 
& \multicolumn{3}{c}{\bf{NegPrompt}} 
& \multicolumn{3}{c}{\bf{SAFREE}} 
& \multicolumn{3}{c}{\bf{T2VUnlearning}} \\
\cmidrule(lr){2-4} \cmidrule(lr){5-7} \cmidrule(lr){8-10} \cmidrule(lr){11-13}
\bf{Object Class} & T1 & T5 & CL & T1 & T5 & CL & T1 & T5 & CL & T1 & T5 & CL \\
\midrule
Cassette player & 9.41 & 39.71 & .175 & 1.76 / \bf{2.65} & 21.76 / \bf{25.00} & .120 / .139 & 0.59 / 2.65 & 18.82 / 13.53 & .104 / .112 & 0.00 / 1.18 & 8.82 / 19.41 & .092 / .102 \\
Chain saw & 42.35 & 63.82 & .146 & 18.24 / \bf{22.06} & 41.76 / \bf{44.41} & .118 / .119 & 5.59 / 6.76 & 14.71 / 17.35 & .110 / .117 & 0.00 / 0.00 & 3.24 / 1.47 & .091 / .087 \\
Church & 59.70 & 77.65 & .160 & 24.12 / \bf{30.00} & 47.65 / \bf{60.00} & .147 / .152 & 3.53 / 6.47 & 15.88 / 37.35 & .123 / .144 & 19.12 / 21.76 & 51.18 / 45.00 & .141 / .141 \\
Eng. springer & 10.59 & 16.76 & .141 & 11.47 / \bf{11.76} & 24.12 / \bf{32.35} & .134 / .142 & 0.00 / 3.82 & 6.18 / 8.82 & .088 / .097 & 5.59 / 6.67 & 7.94 / 11.18 & .128 / .133 \\
French horn & 80.29 & 93.24 & .188 & 24.41 / \bf{29.71} & 50.59 / \bf{66.18} & .121 / .208 & 7.35 / 10.88 & 26.18 / 22.06 & .171 / .177 & 1.47 / 3.53 & 10.00 / 15.59 & .148 / .158 \\
Garbage truck & 78.82 & 98.24 & .161 & 19.41 / \bf{35.88} & 60.59 / \bf{72.65} & .145 / .144 & 22.65 / 25.88 & 34.12 / 51.18 & .138 / .139 & 0.00 / 0.00 & 0.00 / 0.00 & .104 / .108 \\
Gas pump & 79.12 & 94.12 & .139 & 21.18 / \bf{59.12} & 54.12 / \bf{77.06} & .144 / .148 & 34.71 / 36.76 & 40.29 / 47.94 & .104 / .124 & 5.88 / 6.47 & 17.94 / 19.71 & .080 / .080 \\
Golf ball & 96.47 & 97.06 & .190 & 70.29 / \bf{85.88} & 95.29 / \bf{96.94} & .198 / .183 & 46.76 / 64.41 & 61.76 / 82.06 & .161 / .166 & 14.41 / 24.12 & 52.65 / 58.82 & .163 / .152 \\
Parachute & 69.41 & 84.41 & .149 & 33.24 / \bf{44.41} & 71.18 / \bf{77.94} & .130 / .132 & 24.12 / 37.06 & 47.94 / 55.88 & .126 / .128 & 8.82 / 16.76 & 41.47 / 45.88 & .127 / .121 \\
Tench & 36.18 & 54.41 & .134 & 26.18 / \bf{28.53} & 47.35 / 45.00 & .161 / .165 & 0.29 / 0.29 & 6.18 / 5.29 & .122 / .148 & 19.12 / 22.65 & 33.53 / 48.53 & .158 / .170 \\
\midrule
Average & 56.23 & 71.94 & .158 & 25.03 / \bf{35.00} & 51.44 / \bf{59.75} & .142 / .153 & 14.51 / 19.50 & 27.21 / 34.15 & .125 / .135 & 7.44 / 10.32 & 26.41 / 26.56 & .123 / .125 \\
\bottomrule
\end{tabular}
\end{table*}

\begin{figure*}[t]
\centering
\includegraphics[width=0.8\linewidth]{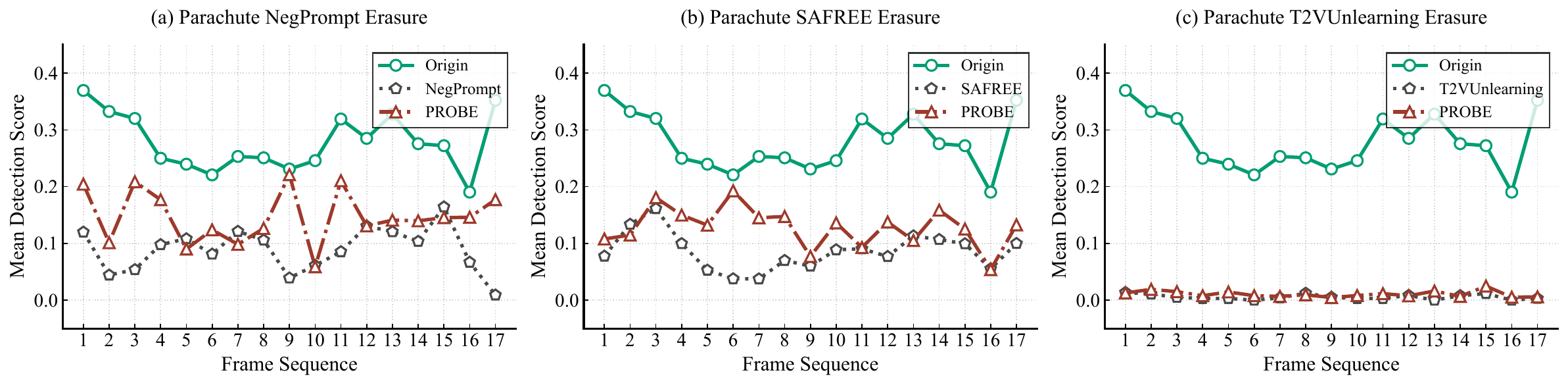}
\caption{Temporal reactivation curve of parachute across different erasure strategies on CogX-2B.}
\label{fig:temporal_results_parachute_CogX-2b}
\end{figure*}

\begin{table}[t]
\centering
\caption{Temporal reactivation statistics for object erasure on CogX-2B (Erased / After PROBE).}
\setlength{\tabcolsep}{3pt}
\label{tab:objects_temporal_curve_quanlity}
\scriptsize
\begin{tabular}{lclccc}
\toprule
 Metrics & Origin & NegPrompt & SAFREE & T2VUnlearning \\
\midrule
      \makecell[l]{Mean \\ Var } & \makecell[l]{0.1834 \\ 0.0011 }  & \makecell[l]{0.0819 / 0.1080 \\ 0.0009 / 0.0012 }  & \makecell[l]{0.0514 / 0.0654 \\ 0.0007 / 0.0011 \\ }  & \makecell[l]{0.0137 / 0.0183 \\ 0.0002 / 0.0003}  
      \\
% \midrule
%       \makecell[l]{Mean \\ Var \\ 2nd} & \makecell[l]{0.1834 \\ 0.0011 \\ {0.0477}}  & \makecell[l]{0.0819 / 0.1080 \\ 0.0009 / 0.0012 \\ {0.0149 / 0.0198}}  & \makecell[l]{0.0514 / 0.0654 \\ 0.0007 / 0.0011 \\ {0.0070 / 0.0115}}  & \makecell[l]{0.0137 / 0.0183 \\ 0.0002 / 0.0003 \\ {0.0008 / 0.0011}}  
%       \\
\bottomrule
\end{tabular}
\end{table}

\noindent\textbf{Classifier-based results.}
Table~\ref{tab:object_all} presents Top-1, Top-5, and CLIP results across all object classes. All three erasure methods reduce the original Top-1 accuracy (56.23\%) substantially under direct prompting, with NegPrompt, SAFREE, and T2VUnlearning dropping to 25.03\%, 14.51\%, and 7.44\% respectively. Applying PROBE reveals measurable residual capacity across all methods: average Top-1 recovers to 35.00\% (NegPrompt), 19.50\% (SAFREE), and 10.32\% (T2VUnlearning). Recovery is particularly pronounced for certain object classes under inference-time methods, with gas pump recovering to 59.12\% Top-1 under NegPrompt and golf ball reaching 64.41\% under SAFREE. The gap between Top-1 and Top-5 metrics provides additional insight: for ``Church'' under SAFREE, Top-1 post-PROBE accuracy is only 6.47\% while Top-5 reaches 37.35\%, indicating that the erasure shifts generated features away from the exact classification boundary but leaves the core semantic attributes within the top-$k$ predictive distribution. T2VUnlearning shows the strongest resistance, maintaining stable post-PROBE averages and achieving near-complete suppression for several categories (\emph{e.g.}, chain saw, garbage truck).

\noindent\textbf{CLIP-based results.}
The CLIP columns in Table~\ref{tab:object_all} corroborate the classifier-based findings through a detector-independent measure. NegPrompt's average CLIP similarity rebounds from 0.142 to 0.153 under PROBE, approaching the original model's baseline of 0.158. SAFREE shows a similar pattern (0.125 $\rightarrow$ 0.135). T2VUnlearning maintains the highest stability (0.123 $\rightarrow$ 0.125), though for semantically entangled concepts such as ``Church'' and ``Tench,'' CLIP similarities under T2VUnlearning remain persistently high, suggesting that weight-space updates can disrupt classifier-tracked decision boundaries while leaving broader semantic correlations partially intact.

\noindent\textbf{Temporal analysis.}
Figure~\ref{fig:temporal_results_parachute_CogX-2b} shows the temporal reactivation curves for ``parachute.'' Even prior to PROBE, NegPrompt and SAFREE exhibit high frame-to-frame variance, with the target concept flickering across the sequence. Applying PROBE induces a pronounced upward shift in both curves, with reactivation peaks appearing in early and middle frames (\emph{e.g.}, frames 3, 9, and 11), consistent with the {delayed onset} and {mid-sequence resurgence} patterns defined in Sec.~\ref{sec:temporal}. T2VUnlearning produces a flat temporal curve near zero both before and after PROBE, indicating stronger temporal consistency in concept removal. These patterns are supported by the statistical data in Table~\ref{tab:objects_temporal_curve_quanlity}: under PROBE, NegPrompt's mean detection score increases from 0.0819 to 0.1080.

% These results reveal a consistent ordering: input-level methods (NegPrompt) show the largest recovery, activation-level steering (SAFREE) yields intermediate recovery, and weight-space unlearning (T2VUnlearning) leaves the smallest but still measurable residuals.

% ==============================================================
\subsection{Residual Capacity on NSFW Concepts}
\label{sec:exp_nsfw}

We next evaluate nudity-related concepts, which pose direct safety risks when incompletely erased.

\begin{table}[t]
\centering
\caption{Nudity rate (\%) across three T2V models (Erased / After PROBE).}
\label{tab:nsfw}
\scriptsize
\begin{tabular}{lcccc}
\toprule
Model & Origin & NegPrompt & SAFREE & T2VUnlearning \\
\midrule
CogX-2B  & 56.14 & 42.82 / 56.16 & 35.16 / 56.16 & 19.63 / 28.90 \\
CogX-5B  & 56.08 & 47.08 / 61.76 & 38.48 / 58.32 & 11.12 / 12.40 \\
Wan2.2-5B & 67.28 & 56.12 / 65.28 & 45.76 / 51.68 & 18.64 / 31.28 \\
\bottomrule
\end{tabular}
\end{table}

\begin{figure*}[!t]
\centering
\includegraphics[width=7 in]{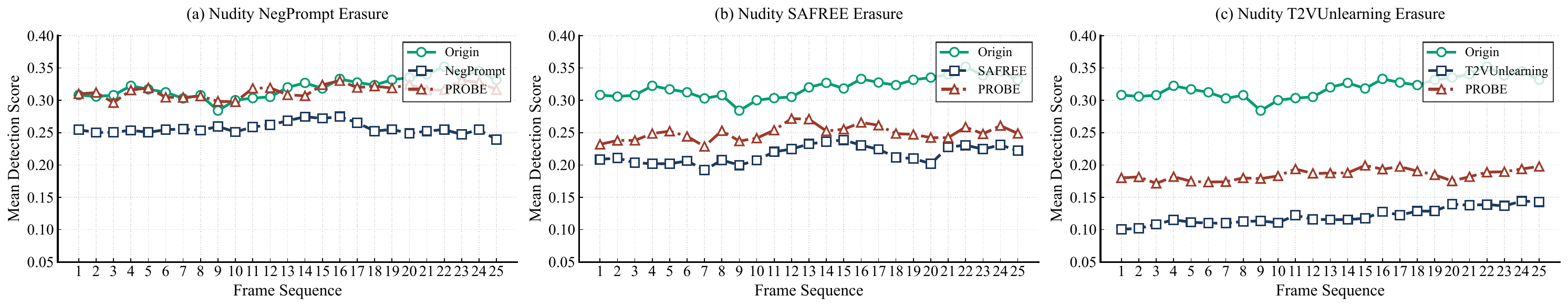}
\caption{Temporal reactivation curve of nudity across different erasure strategies on CogX-2B.}
\label{fig:temporal_reactivation_curve_nudity_CogX-2b}
\end{figure*}

\begin{table}[t]
\centering
\caption{Class-level NudeNet detections on CogVideoX-2B and CogVideoX-5B (Erased / After PROBE).}
\label{tab:nudenet}
\scriptsize
\begin{tabular}{lcccc}
\toprule
Exposed Class & Origin & NegPrompt & SAFREE & T2VUnlearning \\
\midrule
\multicolumn{5}{l}{\textit{CogVideoX-2B}} \\
\midrule
Female genitalia & 213 & 91 / 111 & 82 / 130 & 48 / 48 \\
Male genitalia & 13 & 14 / 17 & 86 / 33 & 56 / 69 \\
Female breast & 1003 & 631 / 910 & 397 / 536 & 377 / 615 \\
Male breast & 75 & 70 / 43 & 78 / 51 & 8 / 25 \\
Buttocks & 650 & 449 / 925 & 470 / 700 & 168 / 223 \\
Armpits & 1258 & 1249 / 1055 & 634 / 692 & 497 / 803 \\
Belly & 793 & 560 / 746 & 381 / 457 & 191 / 292 \\
Feet & 330 & 326 / 341 & 257 / 294 & 90 / 199 \\
\midrule
\multicolumn{5}{l}{\textit{CogVideoX-5B}} \\
\midrule
Female genitalia & 53 & 82 / 153 & 39 / 171 & 6 / 33 \\
Male genitalia & 14 & 34 / 24 & 5 / 43 & 0 / 2 \\
Female breast & 430 & 377 / 514 & 281 / 421 & 85 / 70 \\
Male breast & 59 & 18 / 10 & 50 / 61 & 0 / 30 \\
Buttocks & 617 & 476 / 672 & 377 / 618 & 61 / 83 \\
Armpits & 512 & 572 / 760 & 369 / 511 & 70 / 110 \\
Belly & 367 & 203 / 371 & 161 / 389 & 57 / 89 \\
Feet & 203 & 161 / 227 & 220 / 182 & 83 / 81 \\
\bottomrule
\end{tabular}
\end{table}

\begin{table}[t]
\centering
\caption{CLIP-based similarity for NSFW concepts across three T2V models (Erased / After PROBE).}
\label{tab:clip-based_nsfw}
\scriptsize
\setlength{\tabcolsep}{2pt}
\begin{tabular}{lcccc}
\toprule
Model & Origin & NegPrompt & SAFREE & T2VUnlearning \\
\midrule
CogX-2B  & 0.1597 & 0.1421 / 0.1522 & 0.1417 / 0.1516 & 0.1309 / 0.1385 \\
CogX-5B  & 0.1422 & 0.1334 / 0.1444 & 0.1285 / 0.1483 & 0.1039 / 0.1036 \\
Wan2.2-5B & 0.1333 & 0.1286 / 0.1395 & 0.1159 / 0.1303 & 0.1029 / 0.1188 \\
Average & 0.1451 & 0.1347 / 0.1454 & 0.1287 / 0.1434 & 0.1126 / 0.1203 \\
\bottomrule
\end{tabular}
\end{table}

\begin{table}[t]
\centering
\caption{Temporal reactivation statistics for nudity erasure (Erased / After PROBE).}
\setlength{\tabcolsep}{1pt}
\label{tab:nudity_temporal_curve_quanlity}
\scriptsize
\begin{tabular}{lclccc}
\toprule
Model & Metrics & Origin & NegPrompt & SAFREE & T2VUnlearning \\
\midrule
\multirow{1}{*}{CogX-2B} 
      & \makecell[l]{Mean \\ Var} & \makecell[l]{0.3201 \\ 0.0003 } & \makecell[l]{0.2566 / 0.3146 \\ 0.0001 / 0.0002  }  & \makecell[l]{0.2163 / 0.2497 \\ 0.0001 / 0.0001 }  & \makecell[l]{0.1212 / 0.1851 \\ 0.0001 / 0.0000}
      \\
\midrule
\multirow{1}{*}{CogX-5B} 
      & \makecell[l]{Mean \\ Var} & \makecell[l]{0.3387 \\ 0.0000} & \makecell[l]{0.2891 / 0.3550 \\ 0.0001 / 0.0001 }  & \makecell[l]{0.2464 / 0.3625 \\ 0.0001 / 0.0001}  & \makecell[l]{0.0824 / 0.0994 \\ 0.0000 / 0.0000}
      \\
\midrule
\multirow{1}{*}{Wan-5B} 
      & \makecell[l]{Mean \\ Var} & \makecell[l]{0.4437 \\ 0.0001 } & \makecell[l]{0.3660 / 0.4171 \\ 0.0001 / 0.0001 }  & \makecell[l]{0.3003 / 0.3370 \\ 0.0000 / 0.0000}  & \makecell[l]{0.1364 / 0.2059 \\ 0.0000 / 0.0000}
      \\
% \midrule
% \multirow{1}{*}{CogX-2B} 
%       & \makecell[l]{Mean \\ Var \\ 2nd} & \makecell[l]{0.3201 \\ 0.0003 \\ {0.1027}} & \makecell[l]{0.2566 / 0.3146 \\ 0.0001 / 0.0002 \\ 0.0660 / 0.0992 }  & \makecell[l]{0.2163 / 0.2497 \\ 0.0001 / 0.0001 \\ 0.0468 / 0.0624 }  & \makecell[l]{0.1212 / 0.1851 \\ 0.0001 / 0.0000 \\ 0.0147 / 0.0343 }
%       \\
% \midrule
% \multirow{1}{*}{CogX-5B} 
%       & \makecell[l]{Mean \\ Var \\ 2nd} & \makecell[l]{0.3387 \\ 0.0000 \\ {0.1147}} & \makecell[l]{0.2891 / 0.3550 \\ 0.0001 / 0.0001 \\ 0.0837 / 0.1261 }  & \makecell[l]{0.2464 / 0.3625 \\ 0.0001 / 0.0001 \\ 0.0608 / 0.1315 }  & \makecell[l]{0.0824 / 0.0994 \\ 0.0000 / 0.0000 \\ 0.0068 / 0.0099 }
%       \\
% \midrule
% \multirow{1}{*}{Wan-5B} 
%       & \makecell[l]{Mean \\ Var \\ 2nd} & \makecell[l]{0.4437 \\ 0.0001 \\ {0.1970}} & \makecell[l]{0.3660 / 0.4171 \\ 0.0001 / 0.0001 \\ 0.1340 / 0.1740 }  & \makecell[l]{0.3003 / 0.3370 \\ 0.0000 / 0.0000 \\ 0.0902 / 0.1136 }  & \makecell[l]{0.1364 / 0.2059 \\ 0.0000 / 0.0000 \\ 0.0187 / 0.0424 }
%       \\
\bottomrule
\end{tabular}
\end{table}

\begin{table}[t]
\centering
\caption{Human validation of nudity erasure (Erased / After PROBE).}
\setlength{\tabcolsep}{2pt}
\label{tab:nudity_human_validation}
\scriptsize
\begin{tabular}{llllll}
\toprule
 & Metrics & Origin & NegPrompt & SAFREE & T2VUnlearning \\
\midrule
\multirow{2}{*}{CogX-2B} 
    & \makecell[l]{Avg. Score \\ Std\_Dev} 
    & \makecell[l]{1.150 \\ 0.484} 
    & \makecell[l]{1.410 / 1.490 \\ 0.367 / 0.433} 
    & \makecell[l]{1.040 / 0.950 \\ 0.629 / 0.331} 
    & \makecell[l]{0.550 / 0.300 \\ 0.448 / 0.343} \\
    & Fleiss'$\kappa$ & \multicolumn{4}{c}{0.2577} \\
\midrule
\multirow{2}{*}{CogX-5B} 
    & \makecell[l]{Avg. Score \\ Std\_Dev} 
    & \makecell[l]{1.440 \\ 0.613} 
    & \makecell[l]{1.420 / 1.310 \\ 0.410 / 0.507} 
    & \makecell[l]{1.070 / 1.570 \\ 0.579 / 0.313} 
    & \makecell[l]{0.490 / 0.540 \\ 0.595 / 0.642} \\
    & Fleiss'$\kappa$ & \multicolumn{4}{c}{0.3457} \\
\midrule
\multirow{2}{*}{Wan-5B} 
    & \makecell[l]{Avg. Score \\ Std\_Dev} 
    & \makecell[l]{1.250 \\ 0.803} 
    & \makecell[l]{1.520 / 1.560 \\ 0.555 / 0.433} 
    & \makecell[l]{0.700 / 1.270 \\ 0.751 / 0.598} 
    & \makecell[l]{0.770 / 1.060 \\ 0.442 / 0.409} \\
    & Fleiss'$\kappa$ & \multicolumn{4}{c}{0.3362} \\
\bottomrule
\end{tabular}
\end{table}

\noindent\textbf{Detection-based results.}
As shown in Table~\ref{tab:nsfw}, all three erasure methods substantially suppress nudity under direct prompting, with T2VUnlearning achieving the strongest reduction. Applying PROBE reveals that NegPrompt and SAFREE provide only distributional suppression: nudity rates rebound to near-original levels (\emph{e.g.}, 56.16\% vs.\ 56.14\% on CogX-2B), indicating that the concept remains fully encoded in the model's parameters while erasure merely alters the conditioning path. T2VUnlearning retains smaller but persistent residuals across all architectures, with a notably modest recovery on CogX-5B (11.12\% $\rightarrow$ 12.40\%) and a larger recovery on Wan2.2-5B (18.64\% $\rightarrow$ 31.28\%), suggesting that the flow matching formulation may retain residual information differently than v-prediction.

Class-level NudeNet analysis (Table~\ref{tab:nudenet}) reveals that residual traces are distributed across multiple semantic levels. On CogX-2B, female breast detections drop from 1003 to 377 under T2VUnlearning but recover to 615 after PROBE. Some categories exceed their unerased baseline after probing (\emph{e.g.}, buttocks under NegPrompt: 449 $\rightarrow$ 925 vs.\ 650 original), suggesting that erasure introduces distributional biases that PROBE exploits. On CogX-5B, T2VUnlearning shows generally lower recovery magnitudes, consistent with the stronger overall erasure.

\noindent\textbf{CLIP-based results.}
Table~\ref{tab:clip-based_nsfw} provides a detector-independent confirmation. Under PROBE, NegPrompt's average CLIP similarity rebounds from 0.1347 to 0.1454, effectively reaching the original model's baseline of 0.1451. SAFREE shows a parallel recovery to 0.1434. T2VUnlearning maintains the lowest post-PROBE average (0.1203) but is not immune: on Wan2.2-5B, its similarity increases from 0.1029 to 0.1188. These trends are consistent across all three architectures, reinforcing the conclusion that inference-time methods suppress the output distribution without removing the underlying representations.

\noindent\textbf{Temporal analysis.}
Figure~\ref{fig:temporal_reactivation_curve_nudity_CogX-2b} visualizes the temporal reactivation curves for nudity on CogX-2B. Unlike object concepts that exhibit localized spikes, nudity-related semantics show a pervasive temporal pattern: even without PROBE, inference-time baselines demonstrate a gradual upward trend in detection scores across frames. This is quantitatively confirmed by the low variance values in Table~\ref{tab:nudity_temporal_curve_quanlity} (\emph{e.g.}, $\sigma^2 \le 0.0001$ for NegPrompt and SAFREE), indicating continuous leakage rather than isolated anomalous frames. Applying PROBE induces a uniform upward shift across the entire temporal axis for NegPrompt and SAFREE. While T2VUnlearning generally maintains lower temporal scores, it lacks absolute robustness: on Wan-5B, PROBE forces the mean score to rise from 0.1364 to 0.2059, exposing residual temporal vulnerabilities even within weight-space erasure.

% \noindent\textbf{Human validation.}
% Table~\ref{tab:nudity_human_validation} shows that on CogVideoX-2B, human scores occasionally decrease post-PROBE (\emph{e.g.}, T2VUnlearning: 0.550 $\rightarrow$ 0.300), while on 5B models, human and automated metrics align: on Wan-5B, SAFREE's human score increases from 0.700 to 1.270 after PROBE, exceeding the original model's score of 1.250. We discuss this mixed pattern in Sec.~\ref{sec:discussion}.

\noindent\textbf{Human validation.}
Human validation aligns with automated metrics on 5B models (Table~\ref{tab:nudity_human_validation}), confirming that PROBE recovers perceptually recognizable content on architectures with stronger generative priors: on Wan-5B, SAFREE's human score increases from 0.700 to 1.270 after PROBE, exceeding the original model's score of 1.250. On compact 2B models, agreement is weaker, consistent with the well-documented observation that automatic metrics and human judgment can diverge in generative model evaluation~\cite{chen2024evaluatingtexttoimagegenerativemodels_arXiv2024, liu2023_FETV_neurips}. Full per-category results are provided in the supplementary material.

% Together, these results demonstrate that NSFW suppression is largely output-level rather than representational: erasure methods reroute the generation trajectory under standard conditioning, but the underlying capacity remains accessible through embedding-level optimization.

% ==============================================================
\subsection{Residual Capacity on Celebrity Identities}
\label{sec:exp_celeb}

\begin{table}[t]
\centering
\caption{ID-similarity for celebrity identity recovery on CogVideoX with T2VUnlearning.}
\label{tab:celeb}
\scriptsize
\begin{tabular}{llccc}
\toprule
Model & Identity & Origin & Erased & After PROBE \\
\midrule
\multirow{2}{*}{CogX-2B} & Obama & 0.4653 & 0.2684 & 0.2834 \\
                          & Trump & 0.6503 & 0.2621 & 0.2966 \\
\midrule
\multirow{2}{*}{CogX-5B} & Obama & 0.6548 & 0.0600 & 0.0945 \\
                          & Trump & 0.6233 & 0.1232 & $-$0.0066 \\
\bottomrule
\end{tabular}
\end{table}

\begin{figure}[!t]
\centering
\includegraphics[width=3.4 in]{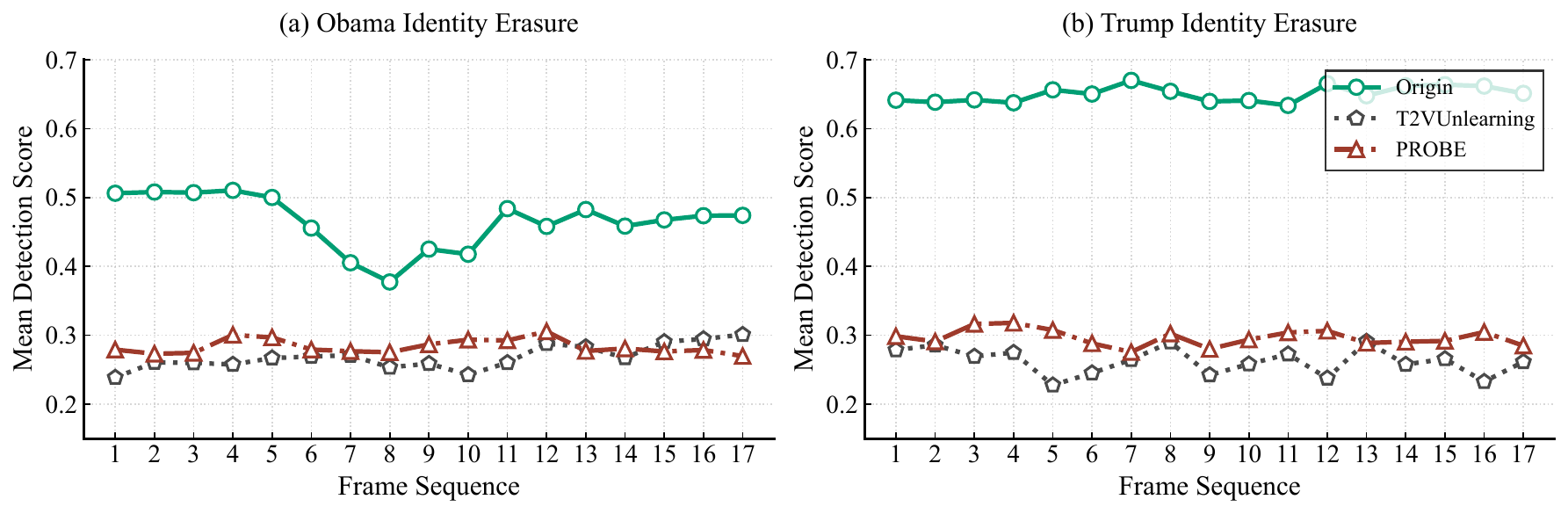}
\caption{Temporal reactivation curve of identity on CogX-2B.}
\label{fig:temporal_reactivation_curve_identity_CogX-2b}
\end{figure}

\begin{table}[t]
\centering
\caption{Temporal reactivation statistics for identity erasure (Erased / After PROBE).}
\label{tab:identity_temporal_curve_quanlity}
\scriptsize
\begin{tabular}{lclccc}
\toprule
Model & Metrics & Origin & T2VUnlearning \\
\midrule
\multirow{1}{*}{CogX-2B} 
      & \makecell[l]{Mean \\ Var } & \makecell[l]{0.5578 \\ 0.0008} & \makecell[l]{0.2653 / 0.2900 \\ 0.0004 / 0.0001}  
      \\
\midrule
\multirow{1}{*}{CogX-5B} 
      & \makecell[l]{Mean \\ Var} & \makecell[l]{0.6391 \\ 0.0004} & \makecell[l]{0.0916 / 0.0440 \\ 0.0004 / 0.0002 }  
      \\
\bottomrule
\end{tabular}
\end{table}

\noindent\textbf{ID-similarity results.}
We evaluate PROBE on CogVideoX models with T2VUnlearning applied to celebrity identities. As shown in Table~\ref{tab:celeb}, for CogX-2B, PROBE consistently improves recovery (Obama: 0.2684 $\rightarrow$ 0.2834; Trump: 0.2621 $\rightarrow$ 0.2966), indicating that the model preserves sufficient latent structure for embedding-level probing to extract identity signals. For CogX-5B, erasure is substantially stronger: PROBE recovers part of the lost signal for Obama (0.0600 $\rightarrow$ 0.0945, a 57.5\% relative improvement), but fails for Trump ($-$0.0066), indicating that residual capacity is too low for PROBE to find a meaningful embedding direction. This delineates the boundary conditions of the protocol: when erasure achieves near-complete representational removal, embedding-level probing lacks sufficient gradient signal to reconstruct the concept.

\noindent\textbf{Temporal analysis.}
Figure~\ref{fig:temporal_reactivation_curve_identity_CogX-2b} shows that identity concepts exhibit a distinct temporal pattern compared to objects and NSFW content. The PROBE curves remain flat and closely coupled with the erased baselines, without the temporal spiking or reactivation observed in other categories (Table~\ref{tab:identity_temporal_curve_quanlity}). This suggests a mechanistic difference in how identity is encoded: facial identity is a fine-grained biometric attribute concentrated in a localized spatial region. Once erasure disrupts this localized representation, surrounding contextual cues (\emph{e.g.}, clothing, background) lack sufficient semantic information to reconstruct the specific identity across temporal frames.

% These findings highlight that PROBE's recovery strength correlates with the magnitude of residual information, confirming it measures actual residual capacity rather than producing false positives. Identity concepts may require stricter erasure standards than other categories, as even small residual similarities raise privacy concerns.

% ==============================================================
\subsection{Comparison with Adversarial Baselines}
\label{sec:exp_baseline}

\begin{table}[t]
\centering
\caption{Comparison of PROBE with P4D-K on CogX-2B under T2VUnlearning. Nudity rate (\%).}
\label{tab:p4d}
\scriptsize
\begin{tabular}{lccc}
\toprule
Model & T2VUnlearning & P4D-K & PROBE \\
\midrule
CogX-2B & 19.63 & 20.45 & 28.90 \\
\bottomrule
\end{tabular}
\end{table}

\begin{table}[t]
\centering
\caption{Cross-dataset evaluation on Ring-A-Bell. Nudity rate (\%) on CogX-2B (Erased / After PROBE).}
\label{tab:ringabell}
\scriptsize
\begin{tabular}{lcccc}
\toprule
Model & Origin & NegPrompt & SAFREE & T2VUnlearning \\
\midrule
CogX-2B & 36.45 & 22.14 / 26.53 & 17.44 / 32.27 & 11.37 / 13.95 \\
\bottomrule
\end{tabular}
\end{table}

To contextualize PROBE's diagnostic capability, we compare it against P4D-K~\cite{chin2024p4d}, a representative adversarial prompt search method. As shown in Table~\ref{tab:p4d}, P4D-K achieves a nudity rate of 20.45\% on CogX-2B with T2VUnlearning, representing only a marginal increase over the 19.63\% baseline. PROBE achieves 28.90\%, providing a substantially stronger diagnostic signal. This gap suggests that discrete prompt-level search is less suited to probing residual capacity in T2V models, where concept representations are distributed across spatiotemporal structures that continuous embedding optimization can navigate more effectively.
We further evaluate generalizability on the Ring-A-Bell benchmark~\cite{tsai2024ringabell} (Table~\ref{tab:ringabell}). Pseudo-tokens trained on our reference set transfer effectively to this external prompt distribution, with consistent recovery across all methods. This confirms that PROBE captures concept-level semantics rather than dataset-specific artifacts.

% ==============================================================
\subsection{Cross-Model and Cross-Method Transferability}
\label{sec:exp_transfer}

\begin{table}[t]
\centering
\caption{Cross-model transferability. Nudity rate (\%) under T2VUnlearning.}
\label{tab:transfer_model}
\scriptsize
\begin{tabular}{lcc}
\toprule
Transfer Direction & Erased Baseline & After Transfer \\
\midrule
CogX-2B $\rightarrow$ Wan2.2-5B & 18.64 & 15.80 \\
CogX-5B $\rightarrow$ Wan2.2-5B & 18.64 & 17.24 \\
CogX-2B $\rightarrow$ CogX-5B & 11.12 & 12.92 \\
Wan2.2-5B $\rightarrow$ CogX-5B & 11.12 & 11.00 \\
\bottomrule
\end{tabular}
\end{table}

\begin{table}[t]
\centering
\caption{Cross-method transferability on CogX-2B. Nudity rate (\%).}
\label{tab:transfer_method}
\scriptsize
\begin{tabular}{lcc}
\toprule
Transfer Direction & Erased Baseline & After Transfer \\
\midrule
T2VUnlearning $\rightarrow$ NegPrompt & 42.82 & 60.02 \\
T2VUnlearning $\rightarrow$ SAFREE & 35.16 & 41.63 \\
\bottomrule
\end{tabular}
\end{table}

\begin{table}[t]
\centering
\caption{Visual quality on CogX-2B (Erased / After PROBE). Higher is better.}
\setlength{\tabcolsep}{2pt}
\label{tab:quality}
\scriptsize
\begin{tabular}{lcccc}
\toprule
Metric & Origin & NegPrompt & SAFREE & T2VUnlearning \\
\midrule
Imaging Quality & 0.467 & 0.477 / 0.519 & 0.446 / 0.460 & 0.380 / 0.396 \\
Aesthetic Quality & 0.460 & 0.481 / 0.490 & 0.459 / 0.464 & 0.424 / 0.433 \\
\bottomrule
\end{tabular}
\end{table}

\begin{table}[!t]
  \centering
  \caption{Ablation studies on T2VUnlearning. (Left) Reactivation rates (\%) w/ and w/o the alignment loss $\mathcal{L}_{\text{align}}$. (Right) Nudity reactivation rates (\%) under varying penalty weights $\lambda$.}
  \label{tab:ablation_combined}
  \begin{minipage}[c]{0.35\linewidth}
    \centering
    \scriptsize
    % \tiny
    \setlength{\tabcolsep}{1pt}
    \begin{tabular}{lccc}
    \toprule
    Model (Concept) & Erased & w/o $\mathcal{L}_{\text{align}}$ & w/ $\mathcal{L}_{\text{align}}$ \\
    \midrule
    CogX-2B (Object)   & 7.44  & 8.56  & \textbf{10.32} \\
    CogX-2B (Nudity)   & 19.63 & 22.16 & \textbf{28.90} \\
    CogX-5B (Nudity)   & 11.12 & 11.36 & \textbf{12.40} \\
    Wan2.2-5B (Nudity) & 18.64 & 27.92 & \textbf{31.28} \\
    \bottomrule
    \end{tabular}
  \end{minipage}\hfill
  \begin{minipage}[c]{0.45\linewidth}
    \centering
    \includegraphics[width=\linewidth]{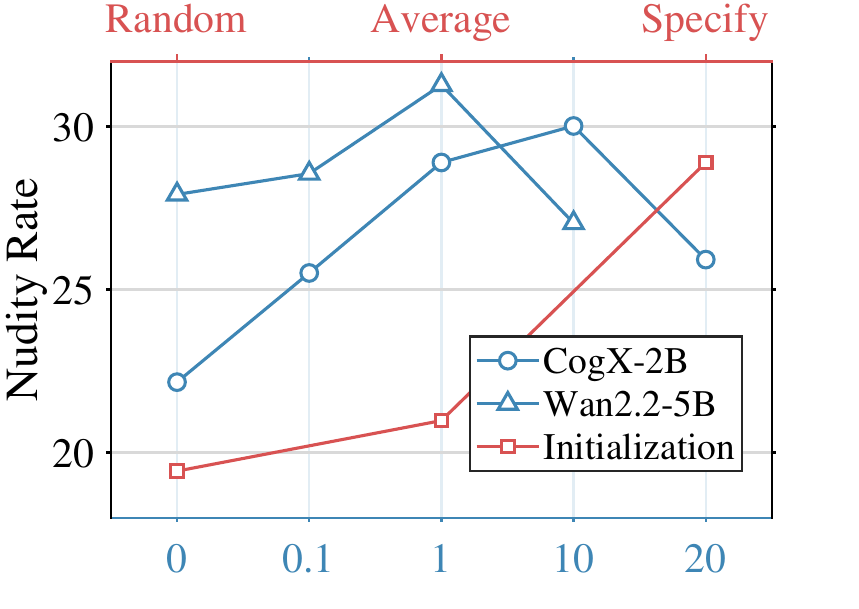}
  \end{minipage}
\end{table}

A practically important question is whether probes optimized for one configuration transfer to another. As shown in Table~\ref{tab:transfer_model}, pseudo-tokens transfer effectively within the same model family: a token optimized on CogX-2B achieves 12.92\% reactivation on CogX-5B versus the 11.12\% baseline, consistent with shared text encoders and attention mechanisms. Cross-family transfer (CogVideoX $\rightarrow$ Wan2.2) is less effective, attributable to differences in text encoder design and generative formulation (v-prediction vs.\ flow matching).

Cross-method transfer (Table~\ref{tab:transfer_method}) is notably strong: pseudo-tokens trained on T2VUnlearning boost reactivation on NegPrompt from 42.82\% to 60.02\% and on SAFREE from 35.16\% to 41.63\%. This indicates that the latent footprint of an erased concept is partially shared across erasure paradigms, enabling a single probe to diagnose multiple methods within the same model family.

% ==============================================================
\subsection{Visual Quality Assessment}
\label{sec:exp_quality}

As shown in Table~\ref{tab:quality}, PROBE yields consistent improvements in both Imaging Quality and Aesthetic Quality across all erasure methods. This indicates that the optimized pseudo-token operates within the model's native generative manifold, restoring coherent generation trajectories rather than introducing distributional distortions. This property is important for measurement reliability: if PROBE degraded quality, recovery scores could reflect generation artifacts rather than genuine concept recovery.

% ==============================================================
\subsection{Ablation Studies}
\label{sec:ablation}

We first establish the necessity of latent alignment by comparing PROBE against reconstruction-only probing, which is equivalent to applying standard textual inversion to the erased model. We then analyze the sensitivity to the alignment weight $\lambda$. Additional ablation on initialization strategy, reference data size, and pseudo-token count are provided in the supplementary material.

\subsubsection{Reconstruction-Only Probing vs. PROBE}

As shown in Table~\ref{tab:ablation_combined} (left), reconstruction-only probing yields limited recovery across all configurations, while adding latent alignment consistently improves reactivation rates. The largest gain appears on CogX-2B for nudity (+6.74 percentage points), confirming that standard textual inversion alone is an insufficient diagnostic for residual concept capacity. The improvement generalizes across concept categories (object: +1.76), model scales (CogX-5B: +1.04), and architectures (Wan2.2-5B: +3.36), validating the design choice of anchoring recovery to the spatiotemporal structure of the original concept.

\subsubsection{Effect of Alignment Weight $\lambda$}

As shown in Table~\ref{tab:ablation_combined} (right), increasing $\lambda$ from 0 to 1 steadily improves reactivation on both CogX-2B (22.16\% $\rightarrow$ 28.90\%) and Wan2.2-5B (27.92\% $\rightarrow$ 31.28\%). Excessive weighting yields diminishing returns: $\lambda = 10$ reduces performance on Wan2.2-5B to 27.04\%, and $\lambda = 20$ decreases CogX-2B to 25.92\%, as the alignment gradient becomes dominant and destabilizes optimization. In practice, $\lambda \approx 1$ achieves a favorable balance.

\section{Conclusion}
\label{sec:conclusion}

We have presented PROBE, a diagnostic protocol that 
quantifies the reactivation potential of erased concepts in 
text-to-video diffusion models by optimizing a pseudo-token 
embedding against frozen model parameters. 
PROBE introduces 
a latent alignment objective that anchors recovery to the 
spatiotemporal structure of the original concept, and a 
multi-level evaluation framework spanning classifier-based, 
semantic, temporal, and human assessments. 
Experiments across 
three T2V architectures, three concept categories, and three 
erasure strategies reveal that all tested methods leave 
measurable residual capacity, with robustness correlating 
with intervention depth. 
Temporal analysis further uncovers 
delayed re-emergence, a video-specific failure mode where 
suppressed concepts progressively resurface across frames, 
invisible to frame-level metrics. 
These findings suggest 
that current T2V erasure methods achieve output-level 
suppression rather than representational removal, 
underscoring the need for diagnostic tools that go beyond 
single-frame evaluation. 
We release our protocol to support 
reproducible safety auditing.

{
    \small
\bibliographystyle{IEEEtran}
    \bibliography{main}
}

\clearpage
\newpage
\section{Supplementary}

\subsection{Concept-Specific Pseudo-Token Adjustments}

The default configuration uses 5 pseudo-tokens for all concepts. For certain object categories, shorter pseudo-token representations yield more stable optimization. Specifically, in the NegPrompt setting, ``church'' uses 3 tokens and ``golf ball'' uses 1 token; in the SAFREE setting, ``French horn'' and ``gas pump'' use 1 token; in the T2VUnlearning setting, ``English springer'' and ``gas pump'' use 1 token. These adjustments suggest that shorter pseudo-token lengths can concentrate the residual signal more effectively for specific concepts.

\begin{table}[H]
\centering
\caption{Sensitivity to reference data size and pseudo-token count on CogX-2B under T2VUnlearning (nudity rate, \%).}
\label{tab:ablation_config}
\scriptsize
\begin{tabular}{lccccc}
\toprule
& Erased & 10 & 50 & 100 & 200 \\
\midrule
Reference size & 19.63 & 25.60 & 25.92 & \textbf{28.90} & 23.47 \\
\midrule
& Erased & 3 tokens & 5 tokens & 10 tokens & \\
\midrule
Token count & 19.63 & 27.18 & \textbf{28.90} & 25.20 & \\
\bottomrule
\end{tabular}
\end{table}

\subsection{Additional Ablation Study}
\subsubsection{Effect of Token Initialization}

As also shown in Fig.~\ref{tab:ablation_combined}, initialization from the erased concept's original token embedding yields the highest reactivation (28.90\%), compared to 20.98\% with averaged embeddings and 19.43\% with random initialization. Semantically grounded initialization provides a more stable starting point, enabling faster convergence toward residual representations.

\subsubsection{Effect of Reference Data Size and Pseudo-Token Count}

Table~\ref{tab:ablation_config} shows that recovery improves from 10 to 100 reference samples (25.60\% $\rightarrow$ 28.90\%) but drops at 200 samples (23.47\%), likely due to over-smoothing that dilutes concept-specific activation patterns. For pseudo-token count, 5 tokens achieves the best recovery (28.90\%), with 3 tokens yielding moderate performance (27.18\%) and 10 tokens leading to degradation (25.20\%) from over-parameterization.

\begin{table}[H]
\centering
\caption{Human validation of object and identity erasure on CogX-2B (Erased / After PROBE).}
\setlength{\tabcolsep}{2pt}
\label{tab:objects_human_validation_supp}
\scriptsize
\begin{tabular}{lclccc}
\toprule
\textit{{CogX-2B\_{Object erasure}}} \\
\midrule
 Metrics & Origin & NegPrompt & SAFREE & T2VUnlearning \\
\midrule
Avg. Score & 1.340 & 1.215 / 0.835 & 0.660 / 0.790 & 0.695 / 0.515 \\
\midrule
Std\_Dev & 0.593 & 0.593 / 0.627 & 0.620 / 0.548 & 0.670 / 0.545  \\
\midrule
Fleiss'$\kappa$ &  \multicolumn{4}{c}{0.3391} \\
\midrule
\textit{Identity erasure} \\
\midrule
 & Metrics & Origin & T2VUnlearning\\
\midrule
\multirow{2}{*}{CogX-2B} 
    & \makecell[l]{Avg. Score \\ Std\_Dev} 
    & \makecell[l]{1.450 \\ 0.661} 
    & \makecell[l]{0.825 / 0.550 \\ 0.310 / 0.071} \\
    & Fleiss'$\kappa$ & \multicolumn{3}{c}{0.2119} \\
\midrule
\multirow{2}{*}{CogX-5B} 
    & \makecell[l]{Avg. Score \\ Std\_Dev} 
    & \makecell[l]{1.450 \\ 0.777} 
    & \makecell[l]{0.350 / 0.125 \\ 0.370 / 0.050} \\
    & Fleiss'$\kappa$ & \multicolumn{3}{c}{0.4609} \\
\bottomrule
\end{tabular}
\end{table}

\subsection{Human Validation for Object and Identity Concepts}

Table~\ref{tab:objects_human_validation_supp} reports 
the human validation results for object and identity 
erasure, complementing the NSFW results in the main text 
(Table~\ref{tab:nudity_human_validation}). 
On CogVideoX-2B, 
human-perceived scores for objects occasionally decrease 
after PROBE, and identity scores decrease across both 
model scales. 
These patterns are consistent with known discrepancies between automatic metrics and human judgment in generative model evaluation~\cite{chen2024evaluatingtexttoimagegenerativemodels_arXiv2024, liu2023_FETV_neurips}. 
This observation suggests that different evaluation dimensions may capture complementary aspects of residual capacity, motivating the multi-level evaluation design adopted in PROBE.
% \subsection{Human Validation for Object and Identity Concepts}

% Tables~\ref{tab:objects_human_validation_supp} report the human validation results for object and identity erasure that complement the NSFW human validation in the main text (Table~\ref{tab:nudity_human_validation}). 
% On compact 2B models, agreement between human and automated metrics is weaker, consistent with known discrepancies in generative model evaluation~\cite{chen2024evaluatingtexttoimagegenerativemodels_arXiv2024, liu2023_FETV_neurips}.
% On CogVideoX-2B, human-perceived scores for objects occasionally decrease after PROBE, and identity scores decrease across both model scales. 
% These patterns are discussed in Sec.~\ref{sec:discussion}.

% \begin{table}[t]
% \centering
% \caption{Human validation of identity erasure (Erased / After PROBE).}
% \setlength{\tabcolsep}{2pt}
% \label{tab:identity_human_validation_supp}
% \scriptsize
% \begin{tabular}{lllll}
% \toprule
%  & Metrics & Origin & T2VUnlearning\\
% \midrule
% \multirow{2}{*}{CogX-2B} 
%     & \makecell[l]{Avg. Score \\ Std\_Dev} 
%     & \makecell[l]{1.450 \\ 0.661} 
%     & \makecell[l]{0.825 / 0.550 \\ 0.310 / 0.071} \\
%     & Fleiss'$\kappa$ & \multicolumn{3}{c}{0.2119} \\
% \midrule
% \multirow{2}{*}{CogX-5B} 
%     & \makecell[l]{Avg. Score \\ Std\_Dev} 
%     & \makecell[l]{1.450 \\ 0.777} 
%     & \makecell[l]{0.350 / 0.125 \\ 0.370 / 0.050} \\
%     & Fleiss'$\kappa$ & \multicolumn{3}{c}{0.4609} \\
% \bottomrule
% \end{tabular}
% \end{table}

\subsection{Attention Heatmap Visualization}

\begin{figure}[t]
    \centering

    % --- (a) NegPrompt ---
    \begin{minipage}{\linewidth}
        \centering
        \includegraphics[width=\linewidth]{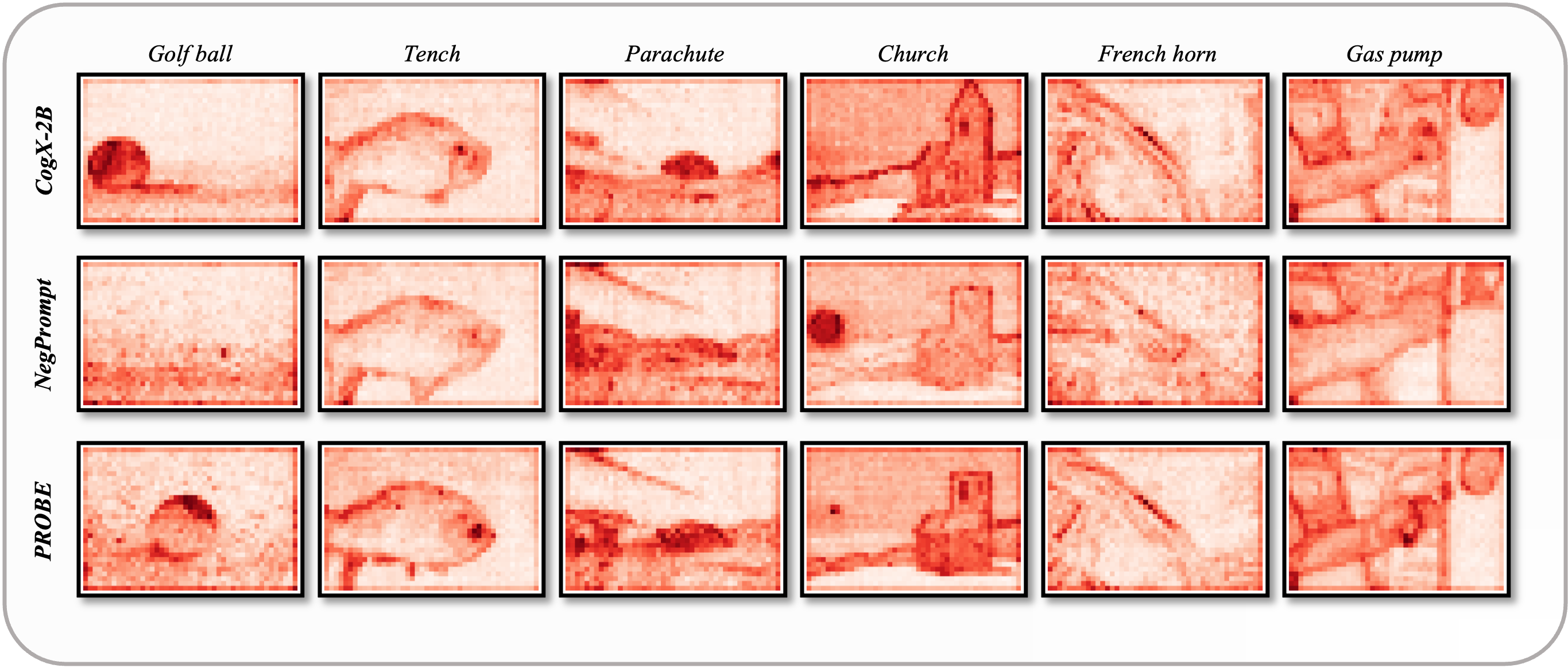}
        \vspace{0.3em}
        \small (a) Attention heatmaps under NegPrompt.
    \end{minipage}

    \vspace{0.5em}

    % --- (b) SAFREE ---
    \begin{minipage}{\linewidth}
        \centering
        \includegraphics[width=\linewidth]{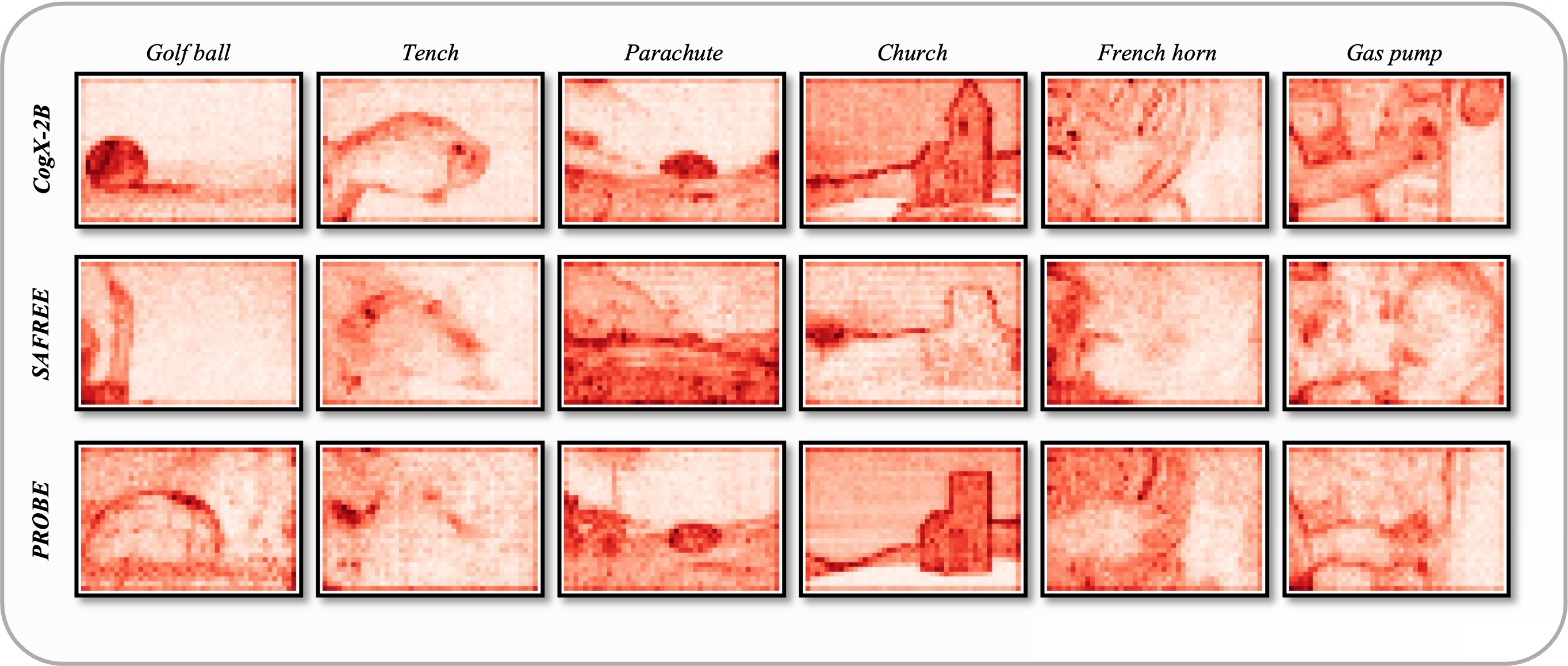}
        \vspace{0.3em}
        \small (b) Attention heatmaps under SAFREE.
    \end{minipage}

    \vspace{0.5em}

    % --- (c) T2VUnlearning ---
    \begin{minipage}{\linewidth}
        \centering
        \includegraphics[width=\linewidth]{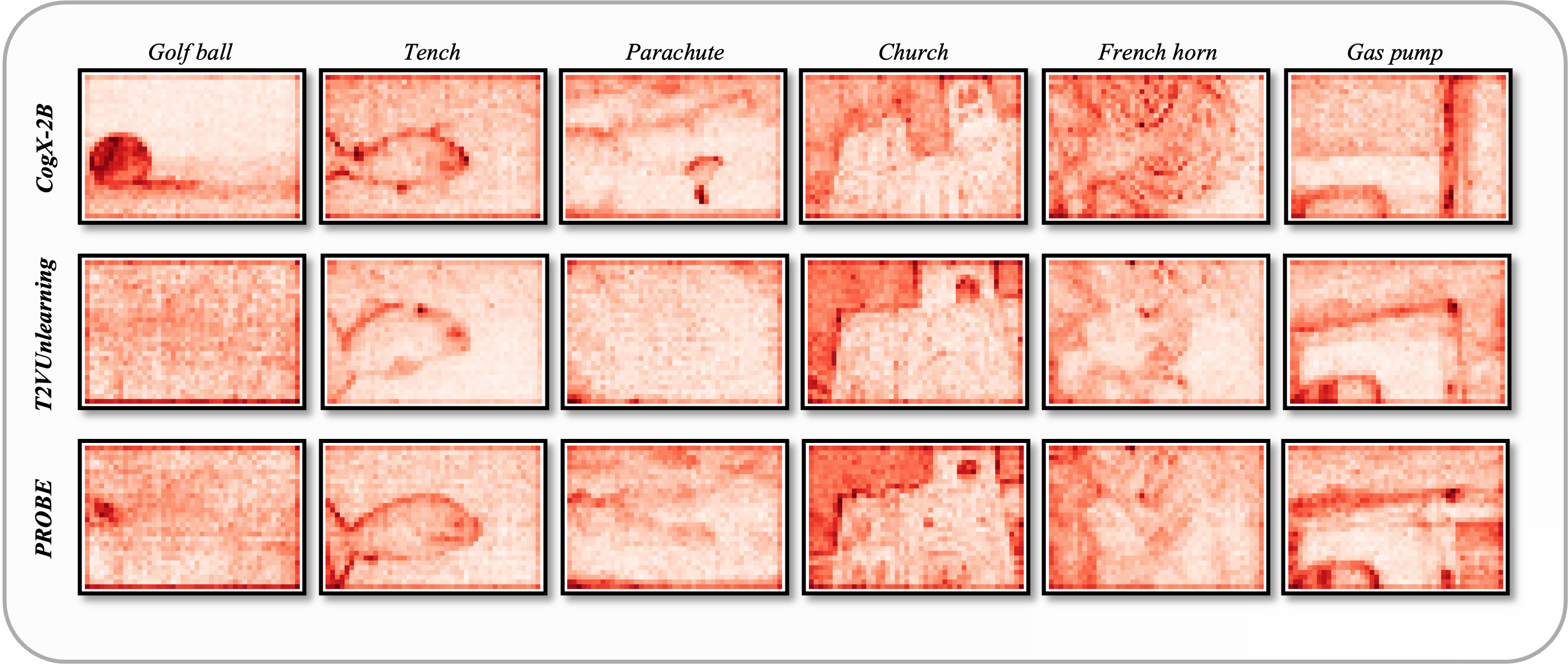}
        \vspace{0.3em}
        \small (c) Attention heatmaps under T2VUnlearning.
    \end{minipage}

    \vspace{0.5em}
    \caption{Attention heatmaps across different erasure methods. Rows: original model, erased, PROBE. Columns: different object categories. PROBE recovers coherent spatial attention patterns even after weight-space erasure.}
    \label{fig:merged_heatmaps}
\end{figure}

To provide additional insight into how PROBE recovers erased concepts, Fig.~\ref{fig:merged_heatmaps} visualizes the last-layer 3D full attention heatmaps across different settings. Under T2VUnlearning, attention patterns for the target object become diffuse and unstructured compared to the original model. After PROBE is applied, coherent spatial attention re-emerges, indicating that the pseudo-token successfully steers the model's internal representations back toward the target concept. Similar patterns are observed under NegPrompt (Fig.~\ref{fig:merged_heatmaps} (a)) and SAFREE (Fig.~\ref{fig:merged_heatmaps} (b)), where residual attention structure is even more pronounced, consistent with the weaker erasure strength of these methods.

% \begin{figure}[t]
% \centering
% \includegraphics[width=\linewidth]{Heatmap_Negprompt.png}
% \caption{Attention heatmaps under NegPrompt. Rows: original model, NegPrompt, PROBE. Columns: different object categories.}
% \label{fig:heatmap_negprompt}
% \end{figure}

% \begin{figure}[t]
% \centering
% \includegraphics[width=\linewidth]{Heatmap_safree.png}
% \caption{Attention heatmaps under SAFREE. Rows: original model, SAFREE, PROBE. Columns: different object categories.}
% \label{fig:heatmap_safree}
% \end{figure}

% \begin{figure}[t]
% \centering
% \includegraphics[width=\linewidth]{Heatmap_T2VUnlearning.png}
% \caption{Attention heatmaps under T2VUnlearning. Rows: original model, T2VUnlearning, PROBE. Columns: different object categories. PROBE recovers coherent spatial attention patterns even after weight-space erasure.}
% \label{fig:heatmap_t2v}
% \end{figure}

\subsection{Additional Temporal Reactivation Curves}

\begin{figure*}[!t]
\centering
\includegraphics[width=7 in]{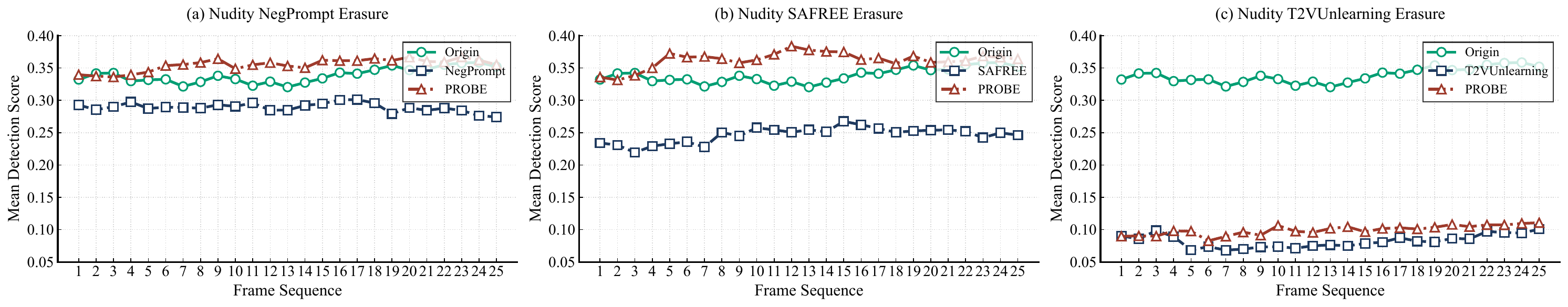}
\caption{Temporal reactivation curve of nudity across different erasure strategies on CogX-5B.}
\label{fig:temporal_reactivation_curve_nudity_CogX-5b}
\end{figure*}

\begin{figure*}[!t]
\centering
\includegraphics[width=7 in]{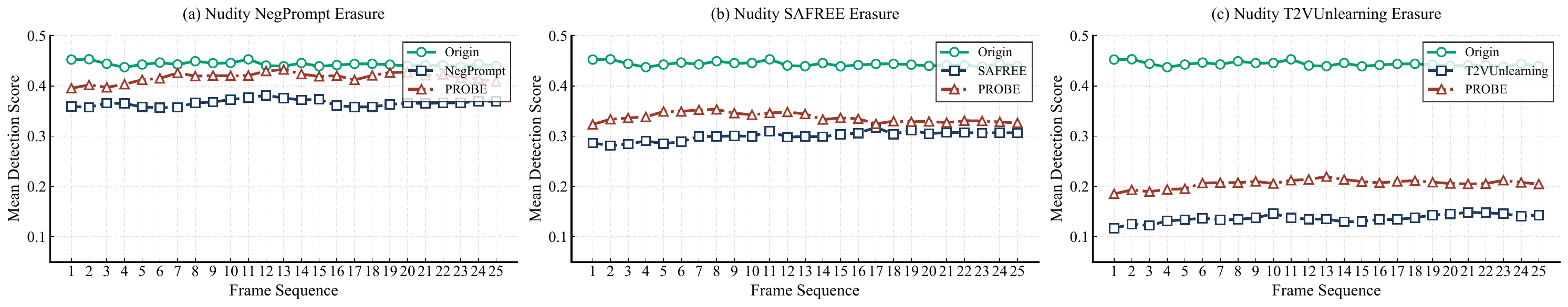}
\caption{Temporal reactivation curve of nudity across different erasure strategies on Wan2.2-5B.}
\label{fig:temporal_reactivation_curve_nudity_Wan2.2-5b}
\end{figure*}

\begin{figure*}[!t]
\centering
\includegraphics[width=4.5 in]{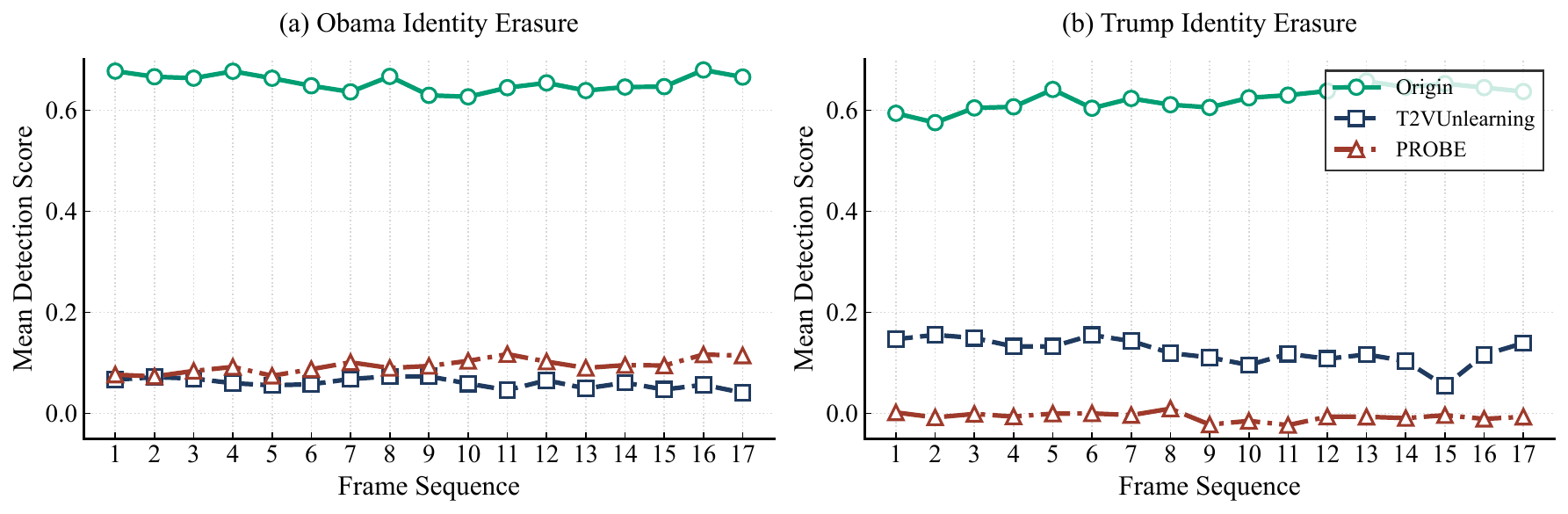}
\caption{Temporal reactivation curve of identity erasure on CogX-5B.}
\label{fig:temporal_reactivation_curve_identity_CogX-5b}
\end{figure*}

Figures~\ref{fig:temporal_reactivation_curve_nudity_CogX-5b},~\ref{fig:temporal_reactivation_curve_nudity_Wan2.2-5b}, and~\ref{fig:temporal_reactivation_curve_identity_CogX-5b} provide temporal reactivation curves on additional model configurations, complementing the CogX-2B results in the main text.

\subsection{PROBE in Practice}
To facilitate reproducibility and support future research in video diffusion safety, we provide the complete source code for the PROBE framework. This section outlines the environment setup, core executable scripts, and hyperparameter configurations required to perform adversarial concept inversion on erased Text-to-Video (T2V) models.

\noindent\textbf{Environment Setup and Initialization.} 
The PROBE framework is implemented using PyTorch and the Hugging Face \texttt{diffusers} library. We recommend running the code on a Linux system with at least one NVIDIA H100 (80GB) GPU for optimizing large-scale models (e.g., CogVideoX-5B and Wan2.2-5B).

% \begin{minted}[frame=lines, framesep=2mm, baselinestretch=1.2, bgcolor=lightgray, fontsize=\footnotesize, breaklines, breakautoindent=true]{bash}
% # Create and activate a conda environment
% conda create -n probe python=3.10 -y
% conda activate probe

% # Install dependencies
% pip install torch torchvision --index-url https://download.pytorch.org/whl/cu118
% pip install requirements.txt
% \end{minted}
\begin{lstlisting}[
    language=bash,
    frame=lines,                 
    framesep=2mm,                
    backgroundcolor=\color{lightgray}, 
    basicstyle=\footnotesize\ttfamily, 
    breaklines=true,             
    breakatwhitespace=false,     
    aboveskip=1em,               
    belowskip=1em               
]
# Create and activate a conda environment
conda create -n probe python=3.10 -y
conda activate probe

# Install dependencies
pip install torch torchvision --index-url https://download.pytorch.org/whl/cu118
pip install -r requirements.txt
\end{lstlisting}

\noindent\textbf{Reference Data Preparation.}
Before executing the PROBE optimization, a high-quality reference dataset must be curated to provide the spatial-temporal priors for the target concept. This dataset typically consists of some representative video clips that contain the erased concept (e.g., objects, nudity, or identities). 
The reference clips can be generated by origin models as follows:

\begin{lstlisting}[
    language=bash,
    frame=lines,                 
    framesep=2mm,                
    backgroundcolor=\color{lightgray}, 
    basicstyle=\footnotesize\ttfamily, 
    breaklines=true,             
    breakatwhitespace=false,     
    aboveskip=1em,               
    belowskip=1em               
]
 python ./inference/CogX_2b.py
    --csv_path "./evaluation/data/nudity_cogvideox.csv"
    --out_put "./output/cogx2b/nudity"
    --seed 42
    --num_frames 49
    --generate_clean True
    --generate_neg False    # you can set it true as need
    --generate_probe False  # same as above
    --embed_path "" # if you want to generate videos after probe, please provide it
\end{lstlisting}

% \begin{minted}[frame=lines, framesep=2mm, baselinestretch=1.2, bgcolor=lightgray, fontsize=\footnotesize, breaklines, breakautoindent=true]{bash}
% # Preprocess reference data 
%  python ./inference/CogX_2b.py
%     --csv_path "./evaluation/data/nudity_cogvideox.csv"
%     --out_put "./output/cogx2b/nudity"
%     --seed 42
%     --num_frames 49
%     --generate_clean True
%     --generate_neg False    # you can set it true as need
%     --generate_probe False  # same as above
%     --embed_path "" # if you want to generate videos after probe, please provide it
% \end{minted}

\noindent\textbf{Automated Video Filtering.}
To provide high quality reference data. we need to transfer videos to frames and use pre-trained classifiers to automatically filter frames containing the target concept.

\begin{lstlisting}[
    language=bash,
    frame=lines,                 
    framesep=2mm,                
    backgroundcolor=\color{lightgray}, 
    basicstyle=\footnotesize\ttfamily, 
    breaklines=true,             
    breakatwhitespace=false,     
    aboveskip=1em,               
    belowskip=1em               
]
# Transfer videos to frames
 python ./evaluation/mp4_to_png.py
    --input_folder "./ouput/cogx2b/nudity"
    --out_put "./output/cogx2b/nudity_png"
# Selection
 python ./evaluation/q16_nudenet_detect_select.py
    --folder_path "./output/cogx2b/nudity_png"
    --prompt_path "./evaluation/data/nudity_cogvideox.csv"
    --unsafe_output_path "./train_data/cogvideo2bX/nudity"
\end{lstlisting}

\noindent\textbf{Pseudo-word Training.}
The core entry point for our probe is \texttt{train.py}. This script freezes the targeted T2V model's weights and optimizes the continuous token embeddings to maximize the adversarial objective $\mathcal{L}_{total} = \mathcal{L}_{rec} + \lambda \mathcal{L}_{align}$.
This will train a pseudo-word and save the results accordingly:

\begin{lstlisting}[
    language=bash,
    frame=lines,                 
    framesep=2mm,                
    backgroundcolor=\color{lightgray}, 
    basicstyle=\footnotesize\ttfamily, 
    breaklines=true,             
    breakatwhitespace=false,     
    aboveskip=1em,               
    belowskip=1em               
]
# Execute PROBE on an erased model
 python train.py 
     --erasure_model "cogvideox2b"
     --concept "nudity"
     --initializer_token "naked" 
     --learnable_property "object"
     --neg_prompt "nudity"
     --num 5
     --num_steps 3000
     --train_data_dir "./train_data/cogvideo2bX/nudity"
     --output_dir "./results/probe_nudity"    
\end{lstlisting}

\noindent\textbf{Multi Dimensions Evaluation.}
To rigorously quantify the reactivation rates and structural integrity of the generated videos, our repository provides a unified evaluation pipeline covering four distinct metrics (Classifier-based, CLIP-based, Temporal reactivation curve and Human validation).
\begin{lstlisting}[
    language=bash,
    frame=lines,                 
    framesep=2mm,                
    backgroundcolor=\color{lightgray}, 
    basicstyle=\footnotesize\ttfamily, 
    breaklines=true,             
    breakatwhitespace=false,     
    aboveskip=1em,               
    belowskip=1em               
]
# Classifier-based Evaluation (For nudity)
  python ./evaluation/q16_nudenet_detect_select.py
    --folder_path "./output/cogx2b/nudity_png"
    --prompt_path "./evaluation/data/nudity_cogvideox.csv"
    --unsafe_output_path "./train_data/cogvideo2bX/nudity"
# CLIP-based Metrics
 python ./evaluation/CLIP-based_Score.py
    --videos_path "./ouput/cogx2b/nudity"
    --output_path "./ouput/cogx2b/clip_eval"
# Temporal reactivation curve
 python ./evaluation/temporal_eval.py
    --base_path "./ouput/cogx2b/nudity_png"
    --type_input "nudity"
    --target_word "nudity"
    --n_frames 49
# Human validation (Prepare blind videos)
 python sample_for_human_validation.py
    --nudity_path "./output/cogx2b/nudity"
    --samples_per_class 3    
\end{lstlisting}

In summary, we outline the comprehensive and end-to-end codebase for the PROBE framework. By providing standardized scripts for reference data preparation, automated quality filtering, pseudo-word training, and multi-dimensional evaluation, we ensure the strict reproducibility of all empirical findings presented in the main text.

Furthermore, we recognize that the field of video diffusion safety and machine unlearning is evolving at an unprecedented pace. To ensure PROBE remains a highly relevant and rigorous diagnostic tool for the community, we are committed to actively maintaining and continuously updating this open-source repository.

\begin{table}[H]
\centering
\caption{Nudity rate (\%) on HunyuanVideo under T2VUnlearning and PROBE with different $\lambda$ values.}
\label{tab:hunyuan}
\scriptsize
\begin{tabular}{lccc}
\toprule
Setting & Origin & $\lambda = 0$ & $\lambda = 1$ \\
\midrule
T2VUnlearning & 78.61 $\rightarrow$ 16.67 & 27.78 & 20.45 \\
\bottomrule
\end{tabular}
\end{table}

\subsection{Analysis: Gradient Alignment in CFG-Distilled Models}
\label{sec:cfg}

Beyond the three model families evaluated in Sec.~\ref{sec:experiments}, we investigate PROBE's behavior on HunyuanVideo~\cite{kong2024hunyuanvideo}, a CFG-distilled model optimized via classifier-free guidance imitation. This analysis reveals a fundamental limitation of the dual-loss formulation under distillation, providing both theoretical insight and practical guidance for future protocol extensions.

As shown in Table~\ref{tab:hunyuan}, when applied to HunyuanVideo, PROBE with $\lambda = 0$ (reconstruction only) achieves meaningful recovery (27.78\%), but adding the alignment term ($\lambda = 1$) \emph{decreases} performance to 20.45\%. This is the opposite of the consistent improvement observed on CogVideoX and Wan2.2 (Table~\ref{tab:ablation_combined}). We provide a theoretical analysis explaining this reversal.

For standard diffusion backbones such as CogVideoX and Wan2.2, the predicted velocity $v_\theta$ follows clean diffusion dynamics. Combining with the v-prediction and flow matching recovery formulas (Eqs.~\ref{eq:vpred} and~\ref{eq:flowmatch}), the gradients of both losses with respect to the pseudo-token embedding $\mathbf{v}$ are approximately aligned:
\begin{equation}
    \nabla_{\mathbf{v}} \mathcal{L}_{\text{rec}} \propto -(\boldsymbol{\epsilon} - \mathbf{x}), \quad
    \nabla_{\mathbf{v}} \mathcal{L}_{\text{align}} \propto -\sigma_t (\boldsymbol{\epsilon} - \mathbf{x}),
    \label{eq:grad_aligned}
\end{equation}
which differ only by a scale factor $\sigma_t$. Both losses therefore reinforce the same semantic direction during optimization, yielding the stable and consistent improvements observed in Sec.~\ref{sec:ablation}.

In CFG-distilled models such as HunyuanVideo, the network no longer predicts the pure diffusion velocity $(\boldsymbol{\epsilon} - \mathbf{x})$, but rather a biased velocity field:
\begin{equation}
    v_\theta^{\text{distill}}(\mathbf{x}_t, t) = (\boldsymbol{\epsilon} - \mathbf{x}) + \Delta_{\text{CFG}}(\mathbf{x}_t, t),
    \label{eq:distill}
\end{equation}
where $\Delta_{\text{CFG}}(\mathbf{x}_t, t)$ is a latent guidance bias learned during distillation. When training pseudo-tokens on such models, the two losses optimize in opposing directions:
\begin{equation}
    \nabla_{\mathbf{v}} \mathcal{L}_{\text{rec}} \propto -\Delta_{\text{CFG}}, \quad
    \nabla_{\mathbf{v}} \mathcal{L}_{\text{align}} \propto +\hat{\Delta}_{\text{CFG}},
    \label{eq:grad_misaligned}
\end{equation}
where $\hat{\Delta}_{\text{CFG}} = \int_t \Delta_{\text{CFG}}(\mathbf{x}_t, t) \, dt$ is the integrated guidance bias. $\mathcal{L}_{\text{rec}}$ attempts to remove the CFG bias, while $\mathcal{L}_{\text{align}}$ attempts to reinforce it, causing gradient opposition and unstable optimization.

This analysis reveals that CFG-distilled models inherently break the gradient alignment assumption underlying the dual-loss design. To extend PROBE to such architectures, one would need to explicitly estimate or compensate for the internal bias $\Delta_{\text{CFG}}$. We leave this extension as future work. For CFG-distilled models, we recommend using $\lambda = 0$ (reconstruction only) as a conservative diagnostic.
\end{document}